\documentclass[10pt,journal,compsoc]{IEEEtran}
\ifCLASSOPTIONcompsoc
  \usepackage[nocompress]{cite}
\else
  \usepackage{cite}
\fi
\ifCLASSINFOpdf
\else
\fi

\usepackage{amsmath,amsfonts}
\usepackage{algorithmic}
\usepackage{algorithm}
\usepackage{array}
\usepackage[caption=false,font=normalsize,labelfont=sf,textfont=sf]{subfig}
\usepackage{textcomp}
\usepackage{stfloats}
\usepackage{url}
\usepackage{verbatim}
\usepackage{graphicx}
\usepackage{cite}
\usepackage{natbib}
\setcitestyle{numbers,square,citesep={,}}
\usepackage{caption}
\usepackage{graphicx}
\usepackage{tcolorbox}
\pdfoutput=1

\usepackage{amsmath}
\usepackage{amssymb}
\usepackage{subfig}
\usepackage{diagbox}
\usepackage{epsfig}
\usepackage{extpfeil}
\usepackage{wrapfig}
\usepackage{multirow}
\usepackage{wrapfig}
\usepackage{color,xcolor,colortbl}
\usepackage[utf8]{inputenc} 
\usepackage[T1]{fontenc}    
\usepackage{url}            
\usepackage{booktabs}       
\usepackage{amsfonts}       
\usepackage{nicefrac}       
\usepackage{microtype}      
\usepackage{xcolor}         
\usepackage{pifont}
\usepackage{xspace}
\usepackage{float}
\usepackage{indentfirst}
\usepackage{tablefootnote}
\usepackage[pagebackref=false,breaklinks=true,colorlinks=true,bookmarks=false,citecolor=blue,linkcolor=red]{hyperref}

\definecolor{lightgray}{gray}{0.95}
\definecolor{color3}{gray}{0.95}
\definecolor{rouse}{rgb}{0.981,0.961,0.941}
\definecolor{light-yellow}{rgb}{1,1,0.93}
\definecolor{light-green}{rgb}{0.95,1,0.95}

\definecolor{color3}{rgb}{0.95,0.95,0.95}
\newcommand{\cmark}{\ding{51}}%
\newcommand{\xmark}{\ding{55}}%

\newcommand{\dataname}{\emph{Motion-X}\xspace}
\newcommand{\newdataname}{\emph{Motion-X++}\xspace}

\begin{document}

%
\title{Motion-X++: A Large-Scale Multimodal 3D Whole-body Human Motion Dataset}
%
%
%
\author{Yuhong Zhang$^{*}$$^{\ddag}$, 
        Jing Lin$^{*}$$^{\ddag}$, 
        Ailing Zeng$^{*}$$^{\ddag}$, 
        Guanlin Wu$^{*}$$^{\ddag}$, 
        Shunlin Lu$^{*}$$^{\ddag}$,  \\
        Yurong Fu,
        Yuanhao Cai,
        Ruimao Zhang~\IEEEmembership{Senior Member,~IEEE} ,
        Haoqian Wang$^{\dag}$, 
        Lei Zhang$^{\dag}$$^{\ddag}$~\IEEEmembership{Fellow,~IEEE} 


\IEEEcompsocitemizethanks{
\IEEEcompsocthanksitem $^{*}$ Equal Contribution, $^{\dag}$ Corresponding Author
\IEEEcompsocthanksitem $^{\ddag}$ Work done at International Digital Economy Academy (IDEA)
\IEEEcompsocthanksitem Yuhong Zhang, Jing Lin, Yurong Fu, and Haoqian Wang are with Shenzhen International Graduate School, Tsinghua University, Guangdong, GD, 518000, China. E-mail: jinglin.stu@gmail.com, \{yuhong-z23, fyr23, wanghaoqian\}@mails.tsinghua.edu.cn
\IEEEcompsocthanksitem Yuhong Zhang, Jing Lin, Ailing Zeng, Guanlin Wu, Shunlin Lu, Lei Zhang are with Department of Computer Vision and Robotics, IDEA, Guangdong, GD, 518000, China. E-mail: \{zhangyuhong, linjing, zengailing, wuguanlin, lushunlin, zhanglei\}@idea.edu.cn. 

\IEEEcompsocthanksitem Guanlin Wu and Yuanhao Cai are with Whiting School of Engineering, Johns Hopkins University, Maryland, MD, 21218, United States. E-mail: gwu32@jh.edu, caiyuanhao1998@gmail.com.

\IEEEcompsocthanksitem Shunlin Lu, Ruimao Zhang are with School of Data Science, The Chinese University of Hong Kong, Shenzhen, Guangdong, GD, 518000. E-mail: shunlinlu0803@gmail.com, ruimao.zhang@ieee.org.

}
}

%
%

\markboth{Journal of \LaTeX\ Class Files,~Vol.~14, No.~8, August~2015}%
{Shell \MakeLowercase{\textit{et al.}}: Bare Demo of IEEEtran.cls for Computer Society Journals}

\IEEEtitleabstractindextext{%

\begin{abstract}
In this paper, we introduce Motion-X++, a large-scale multimodal 3D expressive whole-body human motion dataset.
Existing motion datasets predominantly capture body-only poses, lacking facial expressions, hand gestures, and fine-grained pose descriptions, and are typically limited to lab settings with manually labeled text descriptions, thereby restricting their scalability. To address this issue, we develop a scalable annotation pipeline that can automatically capture 3D whole-body human motion and comprehensive textural labels from RGB videos and build the Motion-X dataset comprising 81.1K text-motion pairs. Furthermore, we extend Motion-X into Motion-X++ by improving the annotation pipeline, introducing more data modalities, and scaling up the data quantities. Motion-X++ provides 19.5M 3D whole-body pose annotations covering 120.5K motion sequences from massive scenes, 80.8K RGB videos, 45.3K audios, 19.5M frame-level whole-body pose descriptions, and 120.5K sequence-level semantic labels.
Comprehensive experiments validate the accuracy of our annotation pipeline and highlight Motion-X++'s significant benefits for generating expressive, precise, and natural motion with paired multimodal labels supporting several downstream tasks, including text-driven whole-body motion generation, audio-driven motion generation, 3D whole-body human mesh recovery, and 2D whole-body keypoints estimation, etc.
\end{abstract}

}

\maketitle

\IEEEdisplaynontitleabstractindextext

\IEEEpeerreviewmaketitle

\IEEEraisesectionheading{\section{Introduction}\label{sec:introduction}}

Human motion studies encompass a range of technologies, emphasizing motion generation and understanding. These advanced methodologies are instrumental in propelling forward various domains, including robotics, embodied systems, animation, gaming, and generative art. Motion generation focuses on creating realistic human movements from specified conditions, allowing users to control and generate motion sequences based on text or audio commands. This approach has gained significant attention due to its high interactivity and intuitiveness~\cite{ahuja2019language2pose,mld,posescript,humanml3d,temos,kit,plappert2018learning,babel,motiondiffuse,modiff}. In parallel, motion understanding, which involves analyzing human behavior for tasks like fine-grained captioning and behavior analysis, is a key component in human-centric multimodal intelligence~\cite{chen2024motionllm,hong2022hcmoco,jiang2024motiongpt,zhou2023avatargpt} and can benefit embodied intelligence from human-computer interaction and robotics to healthcare and security~\cite{wang2022towards, wang2023learning, xiao2024unified}.
%

Despite the substantial contributions of existing text-motion datasets~\cite{humanml3d,amass,kit,babel} to advancing motion generation and motion understanding~\cite{mld,Ho2020DenoisingDP,Song2020DenoisingDI,Tevet2022HumanMD,motiondiffuse}, their scale, diversity, and expressive capability remain unsatisfactory.
Resulting motion from existing datasets~\cite{humanml3d} only includes body movements, neglecting finger movements and facial expressions.
%
The lack of hand gestures and facial expressions impairs the expressiveness and realism of the generated motions.
Besides, certain specialized motions, like advanced skiing, aerial work, and horseback riding, are difficult to capture indoors.
Furthermore, they often rely on single-modal inputs, predominantly text-only or audio-only, which limits the potential applications and versatility of the technology.
In summary, current datasets exhibit four primary limitations: 1) consist solely of body motions without accompanying facial expressions and hand poses; 
2) lack sufficient diversity and are confined to indoor scenes; 
3) encompass a limited range of long-term motion sequences; 
4) depend on unscalable manual text labels, lack professional quality, and are labor-intensive 
and 5) miss multimodal labels and rely solely on single-modal input.
These limitations hinder the ability of existing generation methods to produce expressive, whole-body motions across diverse action types. 
Therefore, \emph{how to collect large-scale whole-body motion and text annotations from multi-scenario videos are critical in addressing the scarcity issue.}

\begin{figure*}[ht]
    \begin{center}
        \begin{tabular}[t]{c} \hspace{-4.6mm}
            \includegraphics[width=0.997\textwidth]{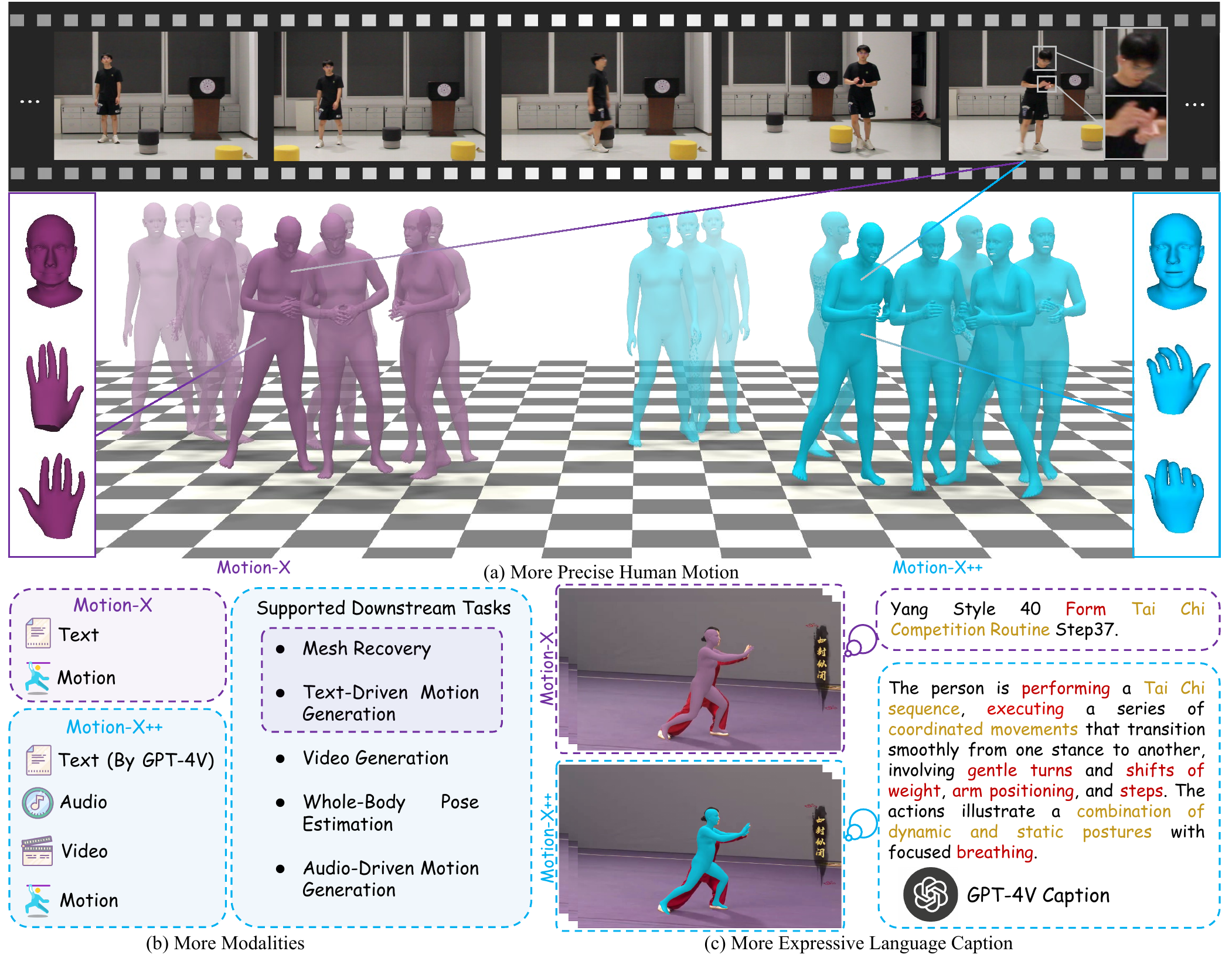}
        \end{tabular}
    \end{center}
    \vspace{-0.27in}
	\caption{\small Compared to \dataname, our enhanced dataset \newdataname offers (a) more precise human motion, including robust facial expressions and refined hand gestures. Facial expressions and hand gestures are highlighted. Additionally, \newdataname provides a broader range of modalities, such as audio and video, and improved quality in text (annotated by GPT-4V) and motion (refined annotation pipeline). The expanded modalities enable \newdataname to support additional downstream tasks, including video generation, whole-body pose estimation, and audio-driven motion generation, beyond mesh recovery and text-driven motion generation supported by \dataname. (c) illustrates a comparison between \dataname and \newdataname, demonstrating more expressive language captions and more precise hand gestures provided by \newdataname.
    }
  \label{fig:motionx_xp_semantic_labels}
  \vspace{-0.25in}
\end{figure*}

Compared to indoor marker-based mocap systems, markerless vision-based motion capture methods~\cite{osx,li2021aist,GyeongsikMoon2020hand4whole,YufeiXu2022ViTPoseSV,glamr,yang2023edpose} show great potential for capturing large-scale motions from extensive video datasets.
Human motion can be conceptualized as a sequence of kinematic structures, which can be automatically converted into pose description using rule-based techniques~\cite{posescript}. 
Although markerless capture methods, such as those utilizing pseudo labels, do not achieve the precision of marker-based approaches, the collection of extensive and informative motions, particularly local motions, remains advantageous~\cite{pang2022benchmarking,osx,moon2022neuralannot,moon2023three,yi2023generating}. 
Besides, text-driven motion generation task requires semantically corresponding motion labels instead of vertex-corresponding mesh labels, and thus have a higher tolerance of motion capture error.
Taking these considerations into account, we propose a scalable and systematic pipeline for motion and text annotation in both multi-view and single-view videos. 

Firstly, we gather and filter massive video recordings from various scenes with challenging, high-quality, multi-style motions and sequence-level semantic labels.
%
Subsequently, we estimate and optimize the parameters of the SMPL-X model~\cite{smpl-x} for comprehensive whole-body motion annotation. 
Due to the depth ambiguity and various scene-specific challenges, existing monocular estimation models typically fail to yield satisfactory results. 
To address this issue, we systematically design a high-performance framework incorporating several innovative techniques.
It includes a hierarchical approach for whole-body keypoint estimation, a score-guided adaptive temporal smoothing and optimization scheme, and a learning-based 3D human model fitting process. 
By integrating these techniques, we can accurately and efficiently capture the ultimate 3D motions.
Finally, we develop an automatic algorithm to generate frame-level descriptions of whole-body poses. 
This involves obtaining body and hand scripts by calculating spatial relations among body parts and hand gestures based on the SMPL-X parameters and extracting facial expressions using an emotion classifier. 
We then aggregate this low-level pose information and translate it into comprehensive textual pose descriptions.

Building on the original annotation pipeline mentioned above, we have developed a large-scale whole-body expressive motion dataset named \dataname, including 15.6M frames and 81.1K sequences with precise 3D whole-body motion annotations, pose descriptions, and semantic labels.
To compile this dataset, we have collected massive videos from the Internet, with a particular focus on game and animation motions, professional performance, and diverse outdoor actions. Additionally, we incorporated data from eight existing action datasets~\cite{cai2022humman,chung2021haa500,liu2019ntu,amass,taheri2020grab,tsuchida2019aist,baum,zhang2022egobody}.
Comprehensive experiments demonstrate its benefits for diverse, expressive, and realistic motion generation. 
%

However, problems such as discontinuous actions from the input video, inaccurate hand gestures, collapsed facial expressions, and imprecise global human trajectory estimation still exist. 
%
To address the aforementioned limitations, we develop an advanced shot detection algorithm capable of processing a broader spectrum of discontinuous video scenes that were previously intractable to existing methodologies and tools. This innovation, coupled with improvements across all motion annotation stages of our established pipeline, we expand the \dataname dataset by incorporating larger-scale data with more diverse scenarios, further expanding the dataset to named \newdataname.

Based on the improved annotation pipeline, \newdataname offers more stable, robust, and accurate global human motions (depicted in Fig.~\ref{fig:motionx_xp_semantic_labels}(a)) than \dataname.
%
%
%
Besides, \newdataname provides more modalities, supporting more downstream tasks, listed in Fig.~\ref{fig:motionx_xp_semantic_labels}(b).
%
%
Fig.~\ref{fig:motionx_xp_semantic_labels}(c) illustrates comparisons between \dataname and \newdataname, highlighting \newdataname's more precise hand gestures and more expressive language caption.
The detailed comparison between \dataname and \newdataname is listed in Tab.~\ref{tab:motionx_xp_compare}.
As one of the largest multimodal human motion datasets, \newdataname contains 19.5M frames and 120.5K sequences with precise 3D whole-body motion annotations with corresponding text, audio, pose descriptions, and annotated whole-body keypoints.
\begin{table*}[t]
	\centering
	\resizebox{1\linewidth}{!}
	{%
            \huge
		\begin{tabular}{l|cc|cccc|cccc}
			\toprule
      \rowcolor{lightgray}
			 & \multicolumn{2}{c|}{\text{Motion Annotation}} &  \multicolumn{4}{c|}{\text{Modalities}}  &  \multicolumn{4}{c}{\text{Supported Downstream Tasks}}\\ 
			\rowcolor{lightgray}
      \multirow{-2}{*}{\text{Dataset}}
      ~& Local Pose Estimation & Global Trajectory Optimization & Text & Motion & Audio & Video & Mesh Rec. & T2M & W. Pose Est. & A2M \\ 
			\midrule
			Motion-X~\cite{lin2023motionx} & EMOCA~\cite{emoca}+OSX~\cite{osx} & GLAMR~\cite{glamr} & \cmark & \cmark & \xmark & \xmark & \cmark & \cmark & \xmark & \xmark \\ 
			Motion-X++ (Ours) & EMOCA{\color{red}$\uparrow$}~\cite{emoca}+HAMER~\cite{hamer}+SMPLer-X~\cite{cai2024smpler} & SLAHMR{\color{red}$\uparrow$}~\cite{ye2023decoupling} & ~~\cmark{\color{red}$\uparrow$} & ~~\cmark{\color{red}$\uparrow$} & \cmark & \cmark & \cmark & \cmark & \cmark & \cmark \\ 
		\bottomrule
	\end{tabular}}
    \vspace{-0.05in}
	\caption{\small Comparison between \dataname and \newdataname. The {\color{red}$\uparrow$} means improved. In local pose estimation stage, \newdataname's annotation pipeline provides refined EMOCA and HAMER, addressing collapsing facial expressions and inaccurate hand gestures. In global optimization stage, we improve SLAHMR by applying mask operation for the camera trajectory estimation and reconstructing foot velocity loss function for global human trajectory estimation. 
    \newdataname provides more modalities and supports more downstream tasks than \dataname. Mesh Rec. means Mesh Recovery, Video Gen. means video generation, and W. Pose Est. means Whole-body Pose Estimation. T2M and A2M mean Text- and Audio-driven motion generation, respectively.
    }
	\label{tab:motionx_xp_compare}
    \vspace{-0.25in}
\end{table*}
We validate \newdataname on various downstream tasks, including music-to-dance, text-to-motion, and 2D whole-body pose estimation. 
With \newdataname, we establish a benchmark for evaluating several SOTA motion generation methods. 
%
%


Our contributions can be summarized as follows:
\begin{itemize}
\item We introduce a large-scale multimodal dataset with precise 3D whole-body motion annotations and corresponding sequence- and frame-level text descriptions.

\item We design an automatic motion and text annotation pipeline, enabling efficient capture of high-quality human text-motion data at scale.

\item \dataname is further extended as \newdataname with more robust and accurate automatic motion annotations with better quality captions, more modalities, and support more downstream tasks.

\item We validate the precision of our motion annotation pipeline and demonstrate the efficacy of \dataname and \newdataname in various downstream tasks, including 3D whole-body motion generation and mesh recovery, through comprehensive experiments.
%
\end{itemize}

An early version of this work~\cite{lin2023motionx} has been published as a conference paper.
We have made significant extensions to our previous work from three aspects.
Firstly, the automatic annotation pipeline of \dataname is improved to be more robust and accurate in both local pose estimation stage, including facial expression estimation, hand gesture estimation, 2D\&3D keypoints estimation, and global trajectory optimization stages. 
Secondly, \dataname is expanded with more datasets such as \textit{UBC, singer, dance} dataset, more modalities such as audio and video corresponding to motion for multimodal task and better annotation quality including text (annotated by GPT-4V) and motion (with our enhanced annotation pipeline). More specifically, the enhanced motion quality includes 1) facial expressions by our advanced EMOCA; 2) hand gestures by HAMER; 3) body pose initialization by SMPLer\-X; 4) 2D\&3D keypoints estimation by enhanced ViT (vision transformer) body parts models and METRABS trained by us; 5) camera trajectory estimation by our proposed masked DROID-SLAM and 6) human trajectory by SLAHMR with improved optimization strategy. Finally, the extended version of \dataname supports more various downstream tasks and has been validated in tasks such as audio-driven motion generation and 2D whole-body pose estimation through comprehensive experiments.
Based on these updates, our final dataset \newdataname with its enhanced precision, expanded modalities, and broad task applicability can support future research in relevant fields.
\vspace{-0.1in}
\section{Related Work}\label{sec:related_work}


%
%
Human motion generation is a key area of research focused on synthesizing realistic and diverse movements. This involves developing datasets and methodologies to address challenges in capturing the complexity of human motion. Its significance lies in advancing human-machine interaction, enhancing virtual environments, and broadening creative expression. Researchers are continually improving techniques to generate lifelike and sophisticated motions, spanning simple to complex actions, for various applications.

\noindent\textbf{Text-Driven Motion Generation.}
Benchmarks annotated with sequential human motion and text are primarily assembled for three tasks: action recognition~\cite{carreira2019short,chung2021haa500,gu2018ava,liu2019ntu,shahroudy2016ntu,trivedi2021ntu}, human object interaction~\cite{hassan2021stochastic,hassan2019resolving,li2019hake,taheri2020grab,zhang2022egobody,zheng2022gimo}, and motion generation~\cite{humanml3d,guo2020action2motion,amass,kit,babel,yi2023generating}.
Notably, KIT Motion-Language Dataset~\cite{kit} is the first public dataset featured with human motion and language descriptions, facilitating multi-modal motion generation~\cite{ahuja2019language2pose,temos}.
Although several indoor human motion capture (mocap) datasets have been developed~\cite{Gross2001TheCM,Ionescu_2014_hm36,Sigal2010HumanEvaSV,Trumble2017TotalC3}, they remain fragmented.
AMASS~\cite{amass} is noteworthy as it consolidates and standardizes 15 different optical marker-based mocap datasets, creating a large-scale motion dataset through a unified framework and parameterization using SMPL~\cite{smpl}. 
This milestone greatly benefits motion modeling and its downstream tasks.
Additionally, BABEL~\cite{babel} and HumanML3D~\cite{humanml3d} enhance the scope of language labels through crowd-sourced data collection.
BABEL provides either sequence labels or subsequence labels for sequential motion, while HumanML3D collects three text descriptions for each motion clip from different annotators.
These text-motion datasets have catalyzed the rapid development of various motion generation methods, demonstrating significant advantages in producing diverse, realistic, and fine-grained motions~\cite{mld,Tevet2022HumanMD,Yuan2022PhysDiffPH,t2m-gpt,motiondiffuse,modiff}. 

\noindent\textbf{Audio Driven Motion Generation.}
Audio, a crucial modality, plays a significant role in driving motion generation. 
Currently, the most popular 3D choreography dataset is AIST++~\cite{tsuchida2019aist}, which offers 5 hours of body-only dance motion data and the corresponding audio, covering 30 different subjects from 9 views, with precise 3D keypoints annotations obtained through multi-view registration. Another notable dataset, Music2Dance~\cite{music2dance}, contains one hour of dance motion data from two genres (modern and folk dance). FineDance~\cite{li2023finedance} utilizes a motion capture system to record a diverse range of dance movements performed by 27 professional dancers across various music tracks.

However, existing text- and audio-motion datasets still have several limitations, including the lack of facial expressions and hand gestures, insufficient data quantity, limited diversity in motions and scenes, coarse-grained and ambiguous descriptions, and the absence of long sequence motions.  
To bridge the gaps in text-motion datasets, \newdataname provides comprehensive sequence- and frame-level text labels. 
Tab.~\ref{tab:comp} presents quantitative comparisons between \newdataname and other existing datasets.
\begin{table*}[t]
	\centering
	\resizebox{1\linewidth}{!}
	{%
		\begin{tabular}{l|cccc|ccc|ccc}
			\toprule
      \rowcolor{lightgray}
			 & \multicolumn{4}{c|}{\text{Motion Annotation}} &  \multicolumn{3}{c|}{\text{Text Annotation}}  &  \multicolumn{3}{c}{\text{Scene}}\\ 
			\rowcolor{lightgray}
      \multirow{-2}{*}{\text{Dataset}}
      ~ & ~~~Clip~~~ & ~~~Hour~~~ & Whole-body?  &{Source} & ~~~Motion~~~ & ~~~Pose~~~ & Whole-body?&Indoor &Outdoor&RGB \\ 
			\midrule
			KIT-ML'16~\cite{kit} & 3911 & 11.2 & \xmark &{Marker-based MoCap} & 6278 & 0 & \xmark & \cmark &\xmark&\xmark \\ 
			AMASS'19~\cite{amass} & 11265 & 40.0 & \xmark &{Marker-based MoCap} & 0 & 0 & \xmark & \cmark &\xmark&\xmark \\ 
			BABEL'21~\cite{babel} & 13220 & 43.5 & \xmark &{Marker-based MoCap} &91408 & 0 & \xmark & \cmark &\xmark&\xmark \\
                Posescript'22~\cite{posescript}&-&-&\xmark&{Marker-based MoCap} &0& 120k&\xmark&\cmark &\xmark&\xmark\\
			HumanML3D'22~\cite{humanml3d} & 14616 & 28.6 & \xmark &{Marker-based MoCap} & 44970 & 0 & \xmark & \cmark &\xmark&\xmark \\ \midrule
                Motion-X~\cite{lin2023motionx} & 81084 & 144.2 & \cmark &{Pseudo GT \& MoCap} & 81084 & 15.6M & \cmark & \cmark & \cmark&\cmark \\
			Motion-X++ (Ours) & 120462 & 180.9 & \cmark &{Pseudo GT \& MoCap} & 120462 & 19.5M & \cmark & \cmark & \cmark&\cmark \\ \bottomrule
	\end{tabular}}
    \vspace{-0.05in}
	\caption{\small Comparisons between \newdataname and existing text-motion datasets. The first column lists the names and publication years of the datasets. \newdataname offers comprehensive whole-body motion and text annotations for both indoor and outdoor environments. Compared to \dataname, \newdataname contains more clips and higher annotation quality.}
	\label{tab:comp}
    \vspace{-0.27in}
\end{table*}
In addition, we encompass several audio-motion paired datasets from various real-world scenarios to address the limitations of existing audio datasets. \newdataname includes the \textit{Music} dataset, featuring various instrument performances, the \textit{Singer} dataset, featuring singers performing, and a large-scale \textit{Dance} dataset collected from the internet. All data have been annotated using our latest single-view annotation pipeline. Moreover, we develop a multi-view whole-body annotation pipeline to refine the AIST dataset, enhancing it with whole-body motion data. This refinement substantially augments the scale and quality of our audio-motion paired data. By proposing \newdataname, we aim to overcome the mentioned limitations and unlock new avenues for future research.
\section{Motion-X++ Dataset}

\subsection{Overview}

\begin{figure}[htbp]
    \vspace{-0.12in}
    \begin{minipage}{0.5\textwidth}
    \centering
    \resizebox{\linewidth}{!}{
        \makeatletter\def\@captype{table}\makeatother
        \begin{tabular}{lccccc}
            \toprule
                \rowcolor{color3} \text{Data} &~~\text{Clip}~~ & ~\text{Frame}~ & {\text{GT Motion}} & {\text{P-GT Motion}} & \text{Text} \\ \midrule
                AMASS~\cite{humanml3d} & 26.3K & 5.4M & B &{H, F} & S \\ 
                HAA500~\cite{chung2021haa500} & 6.9K & 0.4M & - &{B, H, F}& S \\ 
                AIST~\cite{tsuchida2019aist} & 1.4K & 0.3M & - &{B, H, F} & S \\
                HuMMan~\cite{cai2022humman} & 0.9K & 0.2M & - &{B, H, F} & S \\
                GRAB~\cite{taheri2020grab} & 1.3K & 0.4M & B,H &{F} & S \\ 
                EgoBody~\cite{zhang2022egobody} & 1.0K & 0.4M & B,H &{F} & - \\ 
                BAUM~\cite{baum} & 1.4K & 0.2M & - &{F} & S \\
                UBC Fashion~\cite{UBCdwnet} & 0.5K & 0.2M & - & - & - \\
                IDEA400* & 12.5K & 2.6M & - &{B, H, F} & - \\
                Online Videos* & 68.3K & 9.4M & - &{B, H, F} & -\\ \midrule
                Motion-X & 81.1K & 15.6M & B, H &{B,H,F} & S,P \\
                Motion-X++(ours) & 120.5K & 19.5M & B, H &{B,H,F} & S,P  \\ \bottomrule
            \end{tabular}
        }
        \vspace{-0.1in}
        \captionof{table}{\small Statistics of sub-datasets. B, H, F are body, hand, and face. S and P are semantic and pose texts. P-GT is pseudo ground truth. * denotes that videos are collected by us.}
        \vspace{0.08in}
        \label{tab:stat}
    \end{minipage}
    \hfill
    \begin{minipage}{0.5\textwidth}
    \centering

    \includegraphics[width=1\textwidth]{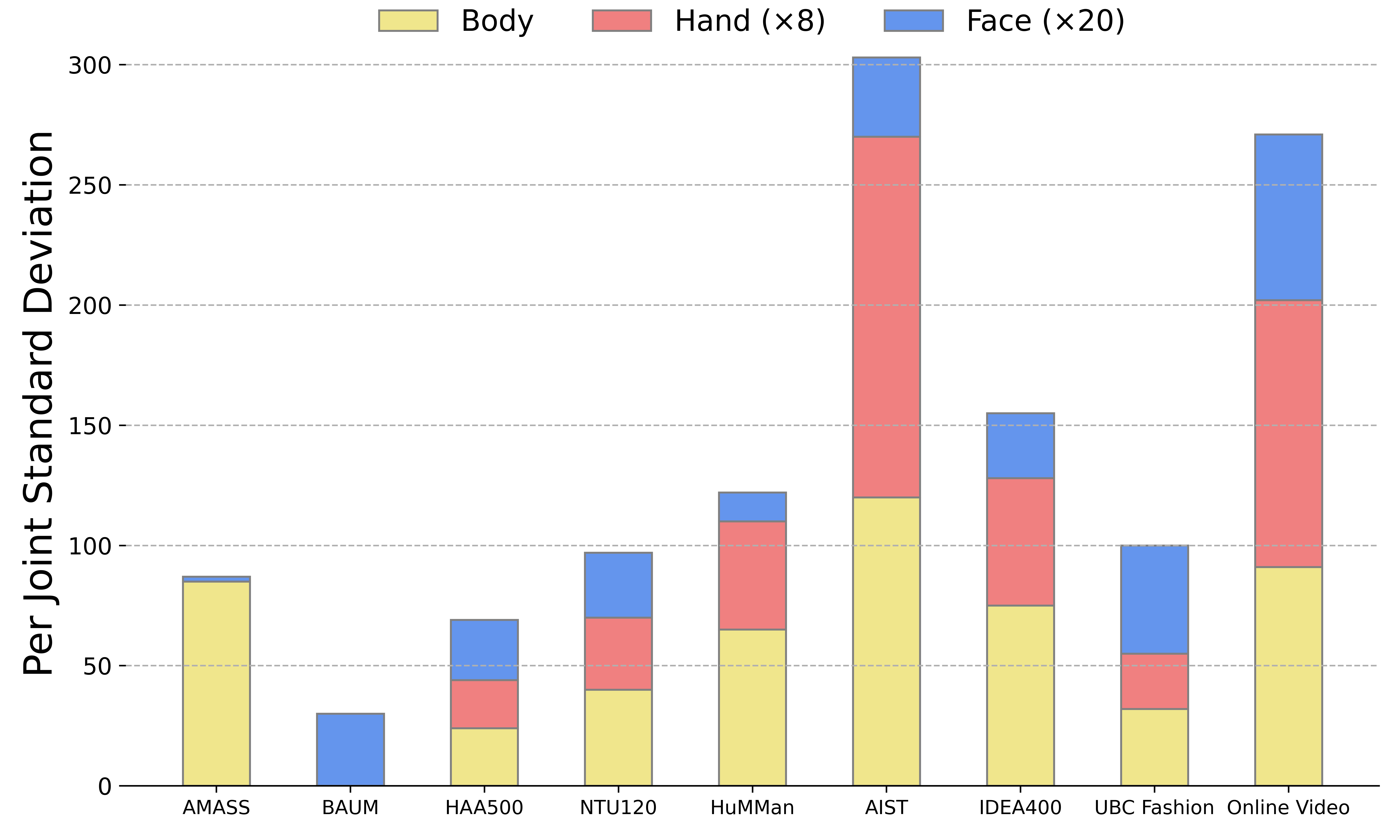}
    \vspace{-0.3in}
    \caption{\small Diversity statistics of the face, hand, and body motions in \newdataname.}
    \label{fig:data_comp} 
    \end{minipage}
    \vspace{-0.1in}
\end{figure}

As shown in Tab.~\ref{tab:stat}, \newdataname is compiled from eight different datasets and online videos, providing
%
19.5M 3D whole-body SMPL-X annotations,
120.5K sequence-level semantic descriptions (e.g., walking with waving hand and laughing), and frame-level whole-body pose descriptions. 
Notably, the original sub-datasets lack either whole-body motion or text labels, which we have unified through our annotation pipeline.
%
To ensure high quality, all annotations have been manually checked.
Fig.~\ref{fig:data_comp} illustrates the average temporal standard deviation of body, hand, and face keypoints for each sub-dataset, highlighting the diversity of hand movements and facial expressions, thereby addressing the gaps present in previous body-only motion data.

\subsection{Data Collection}
\begin{figure*}[t]
    \begin{center}
        \includegraphics[width=1\textwidth]{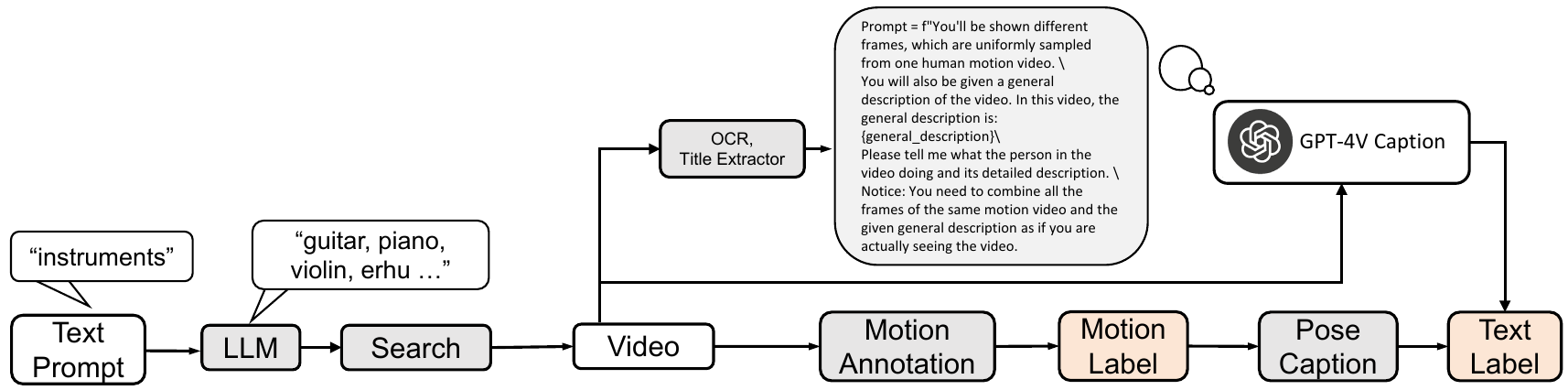}
    \end{center}
    \vspace{-0.2in}
    \caption{\small Illustration of the overall data collection and annotation pipeline.}
    \label{figure:overall_data_collection}
    \vspace{-0.15in}
\end{figure*}

As illustrated in Fig.~\ref{figure:overall_data_collection}, the overall data collection pipeline involves six key steps: 1) designing and sourcing motion text prompts via large language model (LLM)~\cite{jiao2023chatgpt}, 2) collecting videos, 3) preprocessing candidate videos through human detection and video transition detection, 4) capturing whole-body motion (Sec.~\ref{sec:lable_motion}), 5) generating sequence-level semantic label and frame-level whole-body pose description (Sec.~\ref{sec:lable_text}), and 6) conducting manual inspection. 
%

We collect 37K motion sequences from existing datasets using our proposed unified annotation framework.
These datasets include multi-view datasets (AIST~\cite{tsuchida2019aist}), human-scene-interaction datasets (EgoBody~\cite{zhang2022egobody} and GRAB~\cite{taheri2020grab}), single-view action recognition datasets (HAA500~\cite{chung2021haa500} and HuMMan~\cite{cai2022humman}), and body-only motion capture dataset (AMASS~\cite{amass}). 
For these datasets, steps 1 and 2 are omitted. 
Since only the EgoBody and GRAB datasets provide SMPL-X labels with body and hand pose, we annotate the SMPL-X label for other motions. 
For the AMASS dataset, which contains the body and roughly static hand motions, we skip step 4 and augment the facial expression using a data augmentation mechanism. 
The facial expressions are collected from a facial dataset BAUM~\cite{baum} using the EMOCA~\cite{emoca} face capture and animation model.
%
To enhance expressive whole-body motions, we present the IDEA400 dataset, comprising 13K motion sequences across 400 diverse actions. Building on the NTU120~\cite{liu2019ntu} categories, we expand them to include human self-contact, human-object contact, and expressive whole-body motions such as rich facial expressions and detailed hand gestures. The dataset includes 36 actors with varied appearances and clothing, with each action performed ten times: standing(3), walking(3), and sitting(4).
%
%

\begin{figure*}[htbp]
  \begin{center}
      \includegraphics[width=0.99\textwidth]{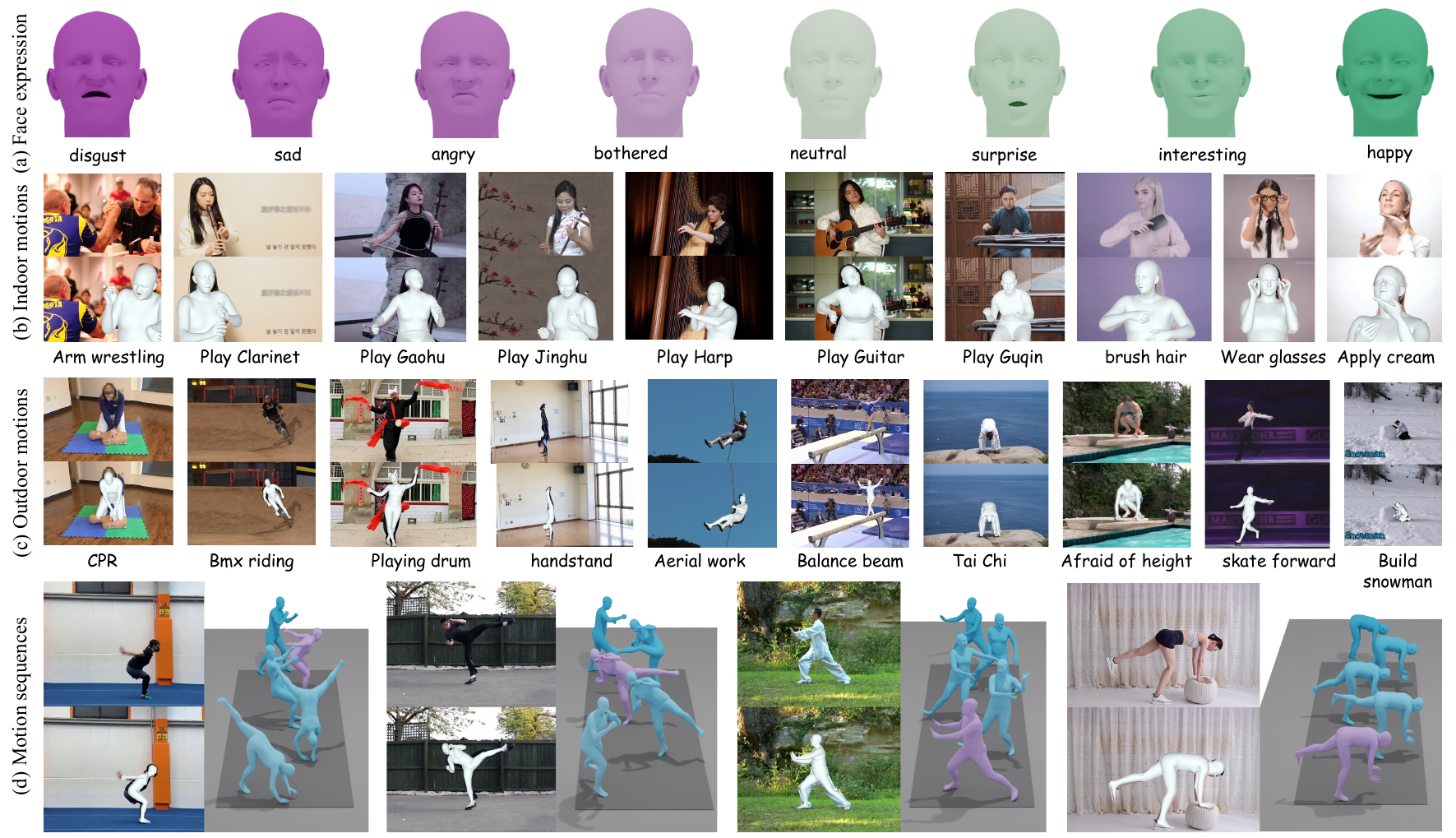}
  \end{center}
  \vspace{-0.2in}
  \caption{\small Overview of \newdataname. It includes (a) diverse facial expressions extracted from BAUM~\cite{baum}, (b) indoor motion with expressive face and hand motions, (c) outdoor motion with diverse and challenging poses, and (d) several motion sequences. Purple SMPL-X is the observed frame, and the others are neighboring poses.}
  \vspace{-0.23in}
  \label{fig:dataset}
\end{figure*}

\noindent\textbf{Diversity.}
To improve the appearance and motion diversity, we collect 32.5K monocular videos from online sources, covering various real-life scenarios as depicted in Fig.~\ref{fig:dataset}.
Recognizing that human motions and actions are context-dependent and vary with the scenario, we design action categories as motion prompts based on context and function of the action via LLM. 
To ensure comprehensive coverage of human actions, our dataset includes general and domain-specific scenes. 
The general scenes encompass daily actions (e.g., brushing hair, wearing glasses, and applying creams), sports activities (e.g., high knee, kick legs, push-ups), musical instrument playing, and outdoor activities (e.g., BMX riding, CPR, building snowman). 
The inclusion of general scenes helps bridge the gap between existing datasets and real-life scenarios.
In addition, we incorporate domain-specific scenes that require high professional skills, such as Kung Fu, Tai Chi, Olympic events, dance, performing arts, entertainment shows, games, and animation motions. 
Based on the prompts describing the above scenes, we utilize the collection pipeline to gather data from online sources for our dataset.


\noindent\textbf{Multimodality.}
Compared to existing text-only or audio-only motion datasets, \newdataname offers multiple modalities, including 120.5K videos with annotated motion sequences, 45.3K audio samples, 19.5M frame-level whole-body pose descriptions, and 120.5K sequence-level semantic labels. This diverse compilation, sourced from multiple data platforms, categorizes sequences by general motion types. To improve semantic density and precision, all text captions are generated using GPT-4V, and synchronized audio is available in music, dance, and singing datasets. Our dataset supports tasks including whole-body pose estimation, mesh recovery, motion synthesis, infilling, prediction, and conditional motion generation. Moreover, it facilitates multimodal unified training and advanced alignment, opening new research opportunities in motion understanding and generation.

\section{Automatic Annotation Pipeline}
To extract precise whole-body motion with continuity, accuracy, and physical plausibility from videos, there are several challenges: 1) multi-shot videos negatively impact motion estimation; 2) incomplete human visibility; 3) large human movements; 4) low resolution or distant camera views causing blurred poses. Our automated annotation pipeline solves these issues with four key stages: 1) Shot Detection segments discontinuous actions using scene detection, tracking, and optical flow; 2) Keypoints Annotation estimates whole-body keypoints with adaptive smoothing; 3) Local Pose Estimation fits the SMPL-X model per frame; 4) Global Trajectory Optimization refines camera and human trajectories. Our proposed pipeline ensures high-quality motion capture from both single- and multi-view inputs.
\label{sec:auto}
\vspace{-0.2in}
\subsection{Shot Detection}
\noindent\textbf{Overview.} Our raw \newdataname videos contain numerous instances of discontinuous action sequences or transitions.
As motion estimation algorithms normally assume that an action captured in an input video is continuous, these issues will compromise the quality of motion data during annotation.
Therefore, we develop a specialized shot detection algorithm that leverages video content analysis, human bounding box tracking, and optical flow to accurately segment video clips with discontinuous actions, known as tracklets.

\noindent\textbf{Tracking and Adaptive Content Shot Detector.}
We start with simple detection of video transitions based on scene detection (SceneDetect) algorithm~\cite{huang2020movienet}.
Then, we design an object tracking-based algorithm based on~\cite{mmtrack2020} to segment action sequences with abrupt positional changes into distinct shots. 
Using Kalman Filter~\cite{kalman1961new}, we identify potentially discontinuous action sequences at the human bounding box granularity. 
When the bounding box position significantly changes between frames, our tracking algorithm treats it as a shot change and assigns different ID to following sequence.

\begin{figure}[H]
    \vspace{-0.1in}
    \centering
    \begin{minipage}[t]{0.16\textwidth}
        \centering
        \includegraphics[width=\textwidth]{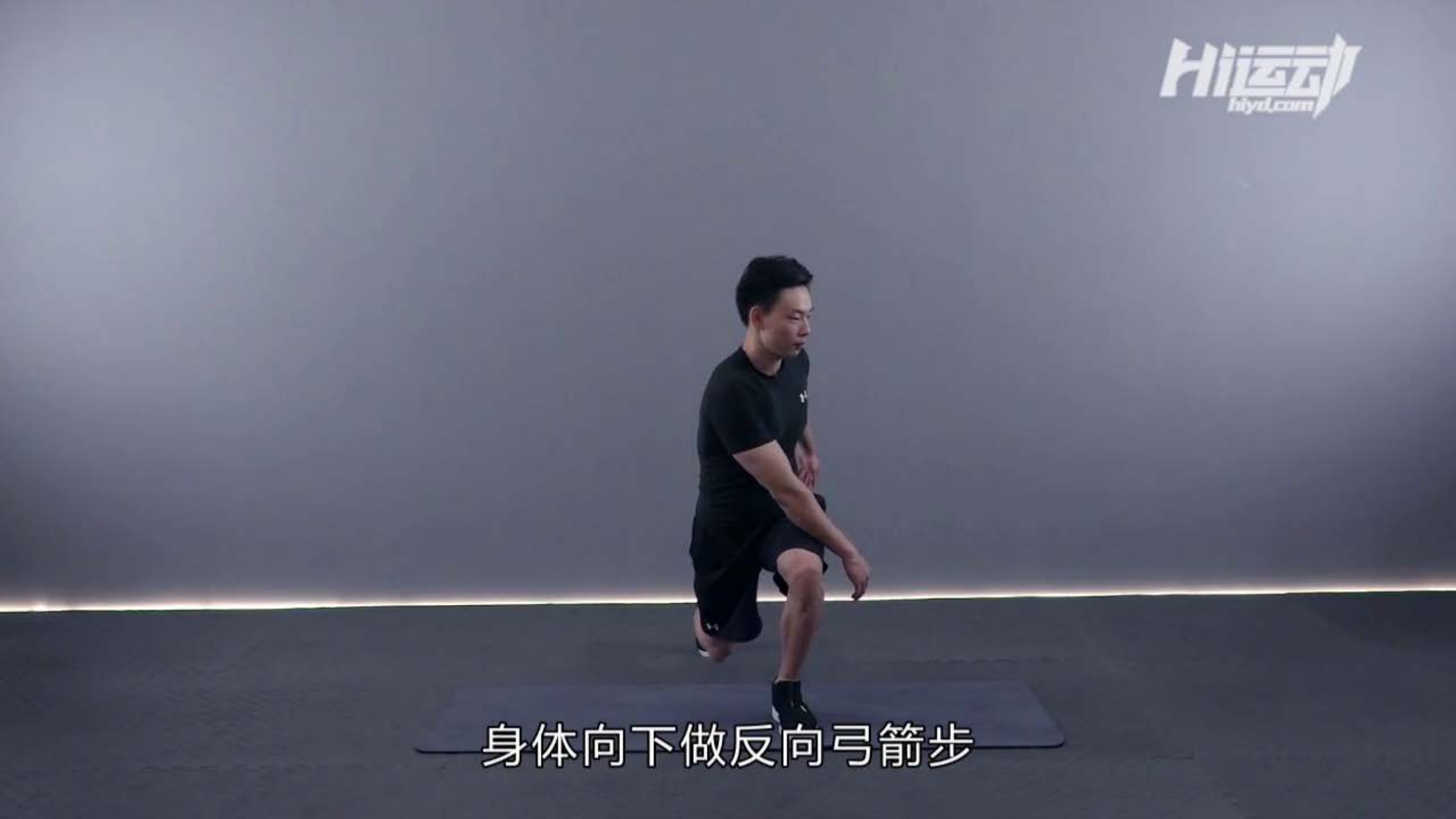}
        \includegraphics[width=\textwidth]{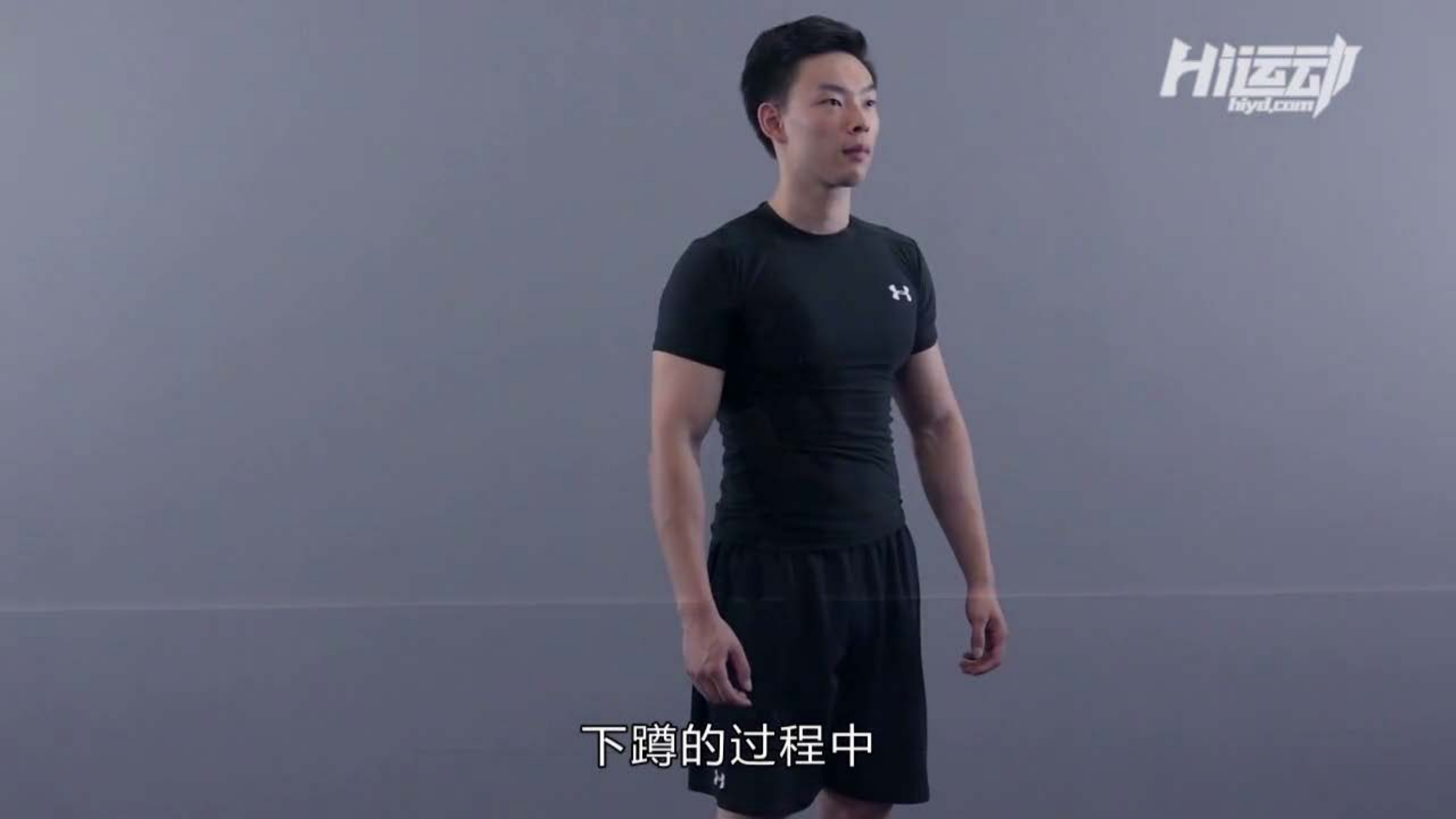}
        \vspace{-0.25in}
        \caption*{\small (a) Scene Change}
    \end{minipage}%
    \hfill
    \begin{minipage}[t]{0.16\textwidth}
        \centering
        \includegraphics[width=\textwidth]{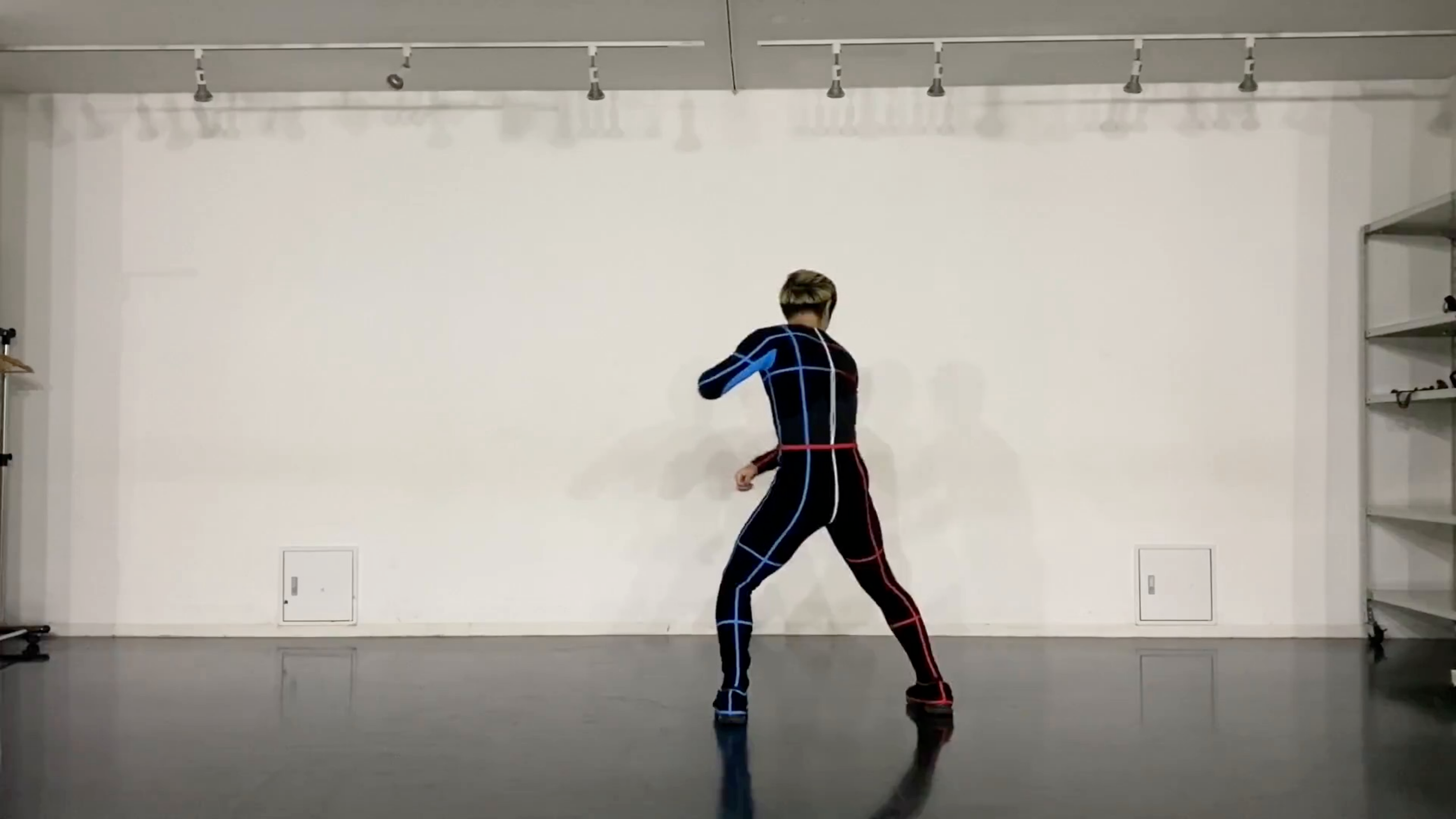}
        \includegraphics[width=\textwidth]{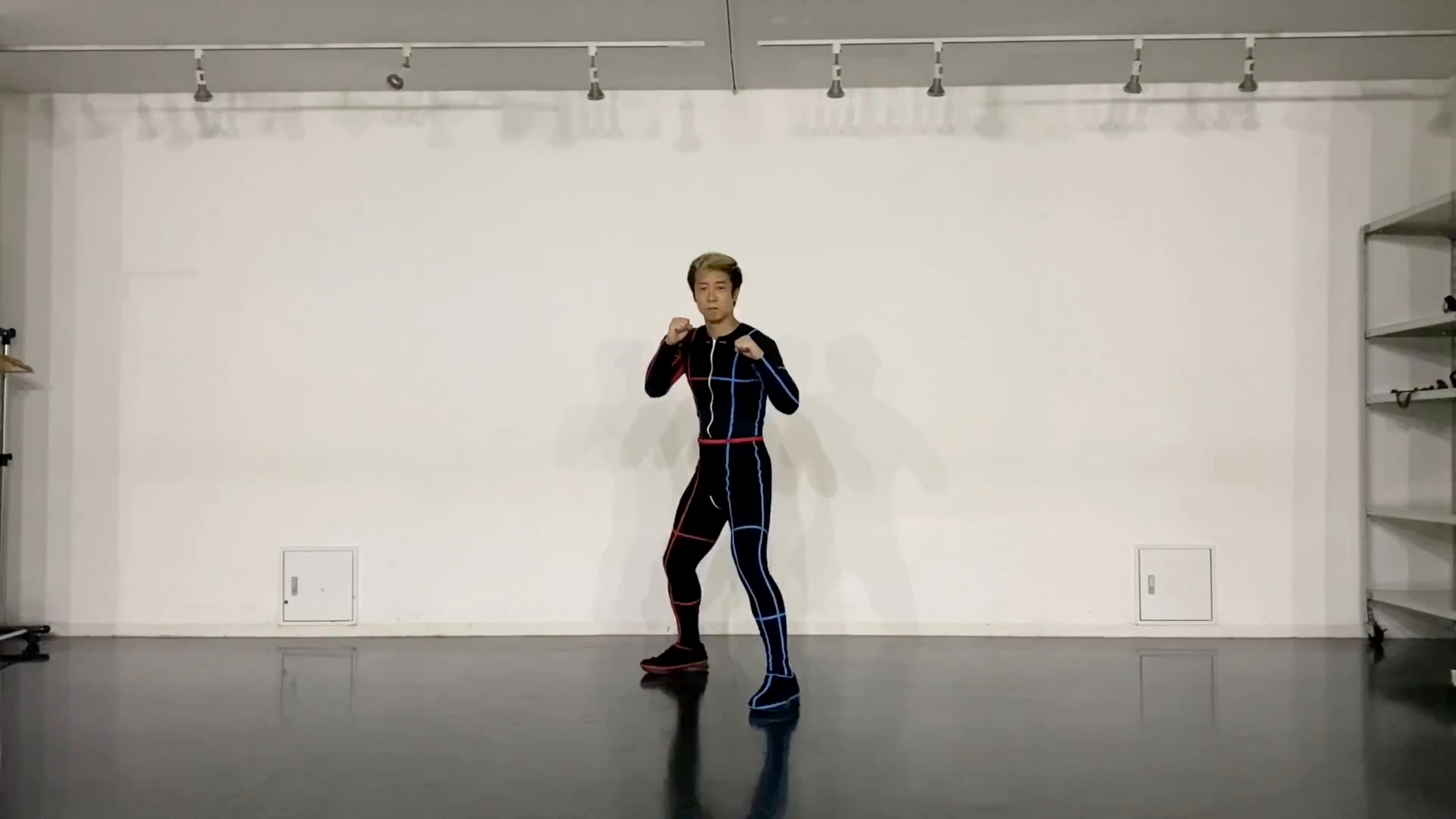}
        \vspace{-0.25in}
        \caption*{\small (b) Position Change}
    \end{minipage}%
    \hfill
    \begin{minipage}[t]{0.16\textwidth}
        \centering
        \includegraphics[width=\textwidth]{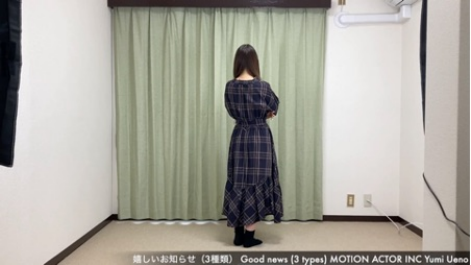}
        \includegraphics[width=\textwidth]{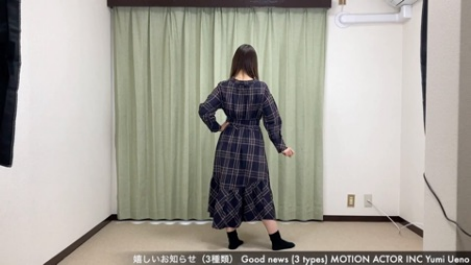}
        \vspace{-0.25in}
        \caption*{\small (c) Pose Change}
    \end{minipage}%
    \captionsetup{font={small,stretch=1}}
    \vspace{-0.1in}
    \caption{\small Examples (a), (b), and (c) illustrate multi-shot scenarios between consecutive frames in our original dataset videos. 
    (a) shows scene changes detectable by SceneDetect. (b) illustrates significant position changes that are undetectable by SceneDetect but resolvable with tracking-based methods. (c) highlights pose changes, which require optical flow-based algorithms for shot detection, as they cannot be addressed by either SceneDetect or tracking-based methods.}
    \label{fig:shot_detection_example}
    \vspace{-0.1in}
\end{figure}

\noindent\textbf{Optical Flow Shot Detector.}
Tracking-based methods cannot handle motion discontinuities below the bounding box granularity, such as changes in direction. 
To address these problems, we design a shot detection algorithm based on optical flow, calculated by RAFT~\cite{teed2020raft}.
This algorithm computes the norm of the mean flow vector inside the bounding box to determine whether there is a shot change between consecutive frames. 
The underlying intuition is that there should not be significant differences between two consecutive frames of human motion, which can be used as a threshold for shot detection.
We provide three examples to illustrate different kinds of multi-shot videos in Fig.~\ref{fig:shot_detection_example}.
%

\subsection{Keypoints Annotation}
\label{sec:lable_motion}

\begin{figure*}[t]
\begin{center}
    \includegraphics[width=1\textwidth]{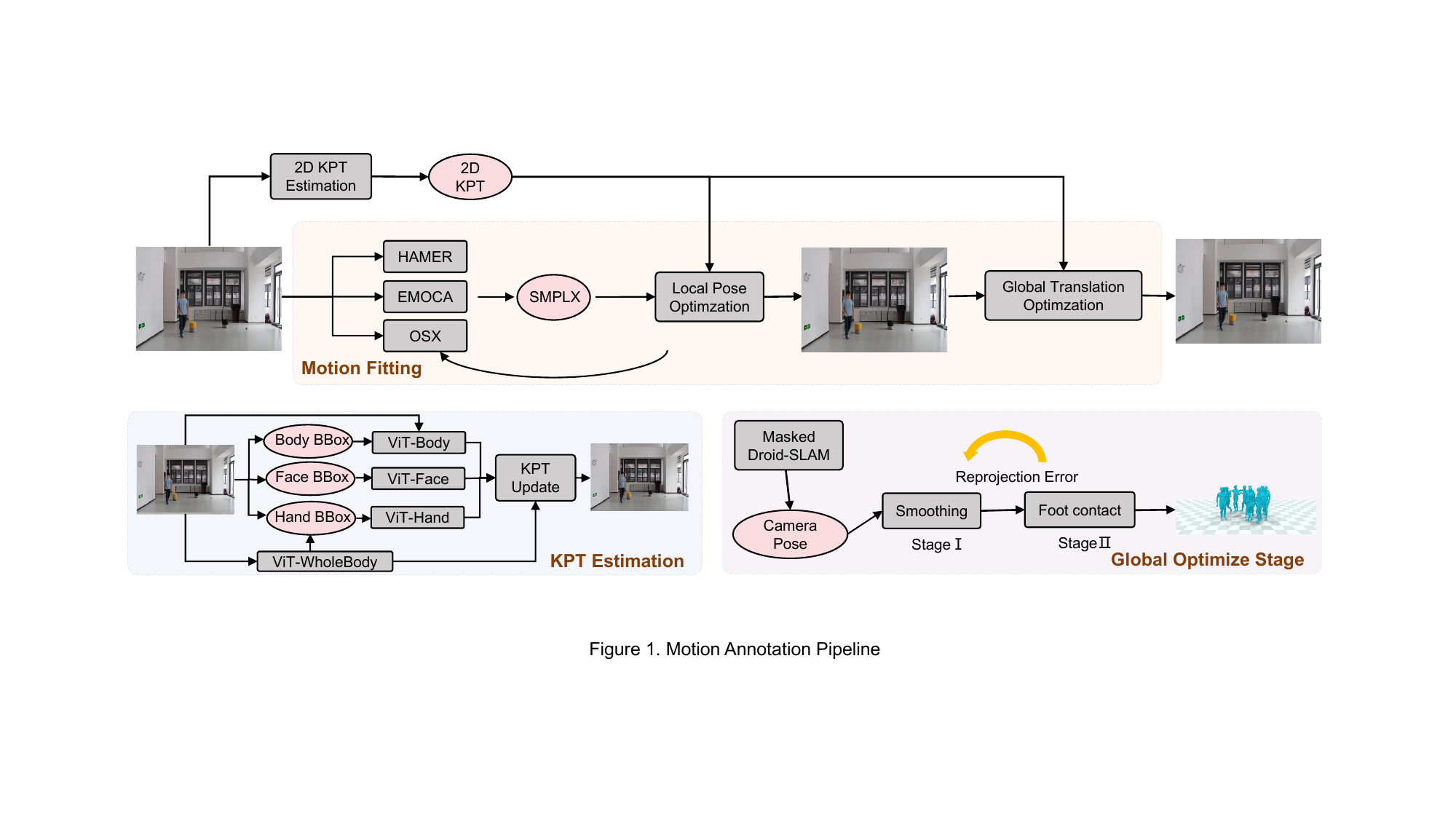}
\end{center}
\vspace{-0.2in}
\caption{\small The automatic pipeline for the whole-body motion capture from massive multi-shot videos. It comprises shot detection, 2D and 3D whole-body keypoints estimation, local pose estimation, and global trajectory optimization stages. This pipeline is designed to support both single-view and multi-view inputs.}

\label{figure:motion_anno}
\vspace{-0.23in}
\end{figure*}

\noindent\textbf{Overview.}
To efficiently capture a large volume of potential motions from massive videos, we propose an annotation pipeline for high-quality whole-body motion capture, incorporating 
\emph{three novel techniques}:
(i) hierarchical whole-body keypoint estimation;
(ii) score-guided adaptive temporal smoothing for jitter motion refinement; 
and (iii) learning-based 3D human model fitting for accurate motion capture. 
\noindent\textbf{2D Whole-body Keypoint Estimation.} 
This task suffers from a significant challenge due to the small size of the hand and face regions.
Although recent approaches have employed separate networks to decode features of different body parts~\cite{coco_wholebody,xu2022zoomnas}, 
they frequently encounter issues such as missed hand detections and errors arising from occlusion or interaction.
To address these limitations, we develop a novel hierarchical keypoint annotation method, illustrated in the blue shaded box of Fig.~\ref{figure:motion_anno}. 
We train a ViT-WholeBody based on a ViT-based model~\cite{YufeiXu2022ViTPoseSV} on the COCO-Wholebody dataset~\cite{coco_wholebody} to estimate initial whole-body 
keypoints $\mathbf{K}^{\text{2D}}\in\mathbb{R}^{133\times2}$ with confidence scores. 
By leveraging the ViT model’s capability to model semantic relationships between full-body parts, we enhance the robustness of hand and face detection, even under severe occlusion.
Subsequently, we derive the hand and face bounding boxes from the keypoints and refine these boxes using the BodyHands detector~\cite{bodyhands} through an IoU matching operation.
Finally, the cropped body, hand, and face regions are fed into three separately pre-trained ViT networks to estimate the keypoints for these regions, which are then used to update $\mathbf{K}^{\text{2D}}$.

\noindent\textbf{Score-guided Adaptive Smoothing.} To mitigate jitter arising from per-frame pose estimation in challenging scenarios such as heavy occlusion, truncation, and motion blur while preserving motion details, we introduce a novel score-guided adaptive smoothing technique into the traditional Savitzky-Golay filter~\cite{savitzky1964smoothing}. 
The filter is applied to a sequence of 2D keypoints of a motion:
\vspace{-0.1in}
\begin{equation}
    \bar{\mathbf{K}}_i^{\text{2D}} = \sum_{j=-w}^w c_j \mathbf{K}^{\text{2D}}_{i+j},
    \label{eq:k_2d}
\end{equation}
where ${\mathbf{K}}_i^{\text{2D}}$ represents the original keypoints of the $i_\text{th}$ frame, $\bar{\mathbf{K}}_i^{\text{2D}}$ denotes the smoothed keypoints, $w$ corresponds to half-width of filter window size, and $c_j$ are the filter coefficients.
Unlike existing smoothing methods that use a fixed window size~\cite{zeng2022deciwatch,zeng2022smoothnet,savitzky1964smoothing}, we leverage the confidence scores of the keypoints to adaptively adjust the window size, balancing between smoothness and motion details. 
Employing a larger window size for keypoints with lower confidence scores helps mitigate the impact of outliers.

\noindent\textbf{3D Keypoint Annotation.} Accurate 3D keypoints can significantly enhance the estimation of SMPL-X.
To achieve this, we leverage novel information derived from large-scale pre-trained models.
 Specifically, for single-view videos, we adopt a pre-trained mode~\cite{sarandi2023learning}, trained on extensive 3D datasets~\cite{h36m_pami, singleshotmultiperson2018,Joo_2017_TPAMI,HuCVPR2019,aist-dance-db}, to estimate precise 3D keypoints.

\vspace{-0.1in}
\subsection{Local Pose Estimation}
After obtaining the keypoints, we perform local pose optimization to register each frame's whole-body model SMPL-X~\cite{smpl-x}. 
Traditional optimization-based methods~\cite{smplify, smpl-x} are often time-consuming and may yield unsatisfactory results as they neglect image clues and motion prior. 
To address these limitations, We propose a progressive learning-based human mesh fitting method. 
Initially, we predict the SMPL-X parameter $\Theta$
using the state-of-the-art whole-body mesh recovery method SMPLer-X~\cite{cai2024smpler}, face reconstruction model EMOCA~\cite{emoca} and hand pose estimate method HAMER~\cite{hamer}. 
More specifically, SMPLer-X estimates the 3D body rotations $\theta_b\in\mathbb{R}^{22\times3}$, body shape parameters $\beta_b\in\mathbb{R}^{10}$. EMOCA predicts the yaw pose $\theta_f\in \mathbb{R}^{3}$, face expression code $\psi\in \mathbb{R}^{3}$ and face shape parameters $\beta_f\in \mathbb{R}^{50}$. Hand rotations $ \theta_h\in\mathbb{R}^{30\times3}$ are predicted by HAMER. Once all the local pose parameters have been estimated, we combine the face, hand, and body parameters together as input for the local fitting stage.
%
%
Subsequently, through iterative optimization of the network parameters, we fit the human model parameters $\hat{\Theta}$ to the target 2D and 3D joint positions by minimizing the Eq.~\ref{eq:ljoint}, getting an improved alignment accuracy:
\begin{equation}
    \vspace{-0.09in}
    L_\text{joint} = \Vert \hat{\mathbf{K}}^\text{3D}-\bar{\mathbf{K}}^\text{3D} \Vert_1 + \Vert \hat{\mathbf{K}}^\text{2D}-\bar{\mathbf{K}}^\text{2D} \Vert_1
     + \Vert \hat{\Theta} - \Theta \Vert_1,
     \label{eq:ljoint}
\end{equation}
where $\bar{\mathbf{K}}^\text{3D}$ means smoothed 3D joint positions estimates from~\cite{sarandi2023learning} and $\hat{\mathbf{K}}^\text{3D}$ represents the predicted 3D joint positions obtained by applying a linear regressor to a 3D mesh generated by the SMPL-X model. 
$\hat{\mathbf{K}}^\text{2D}$ is derived by performing a perspective projection of the 3D joint and $\bar{\mathbf{K}}^\text{2D}$ is calculated from Eq.~\ref{eq:k_2d}.
The last term of the loss function provides explicit supervision based on the initial parameter, serving as a 3D motion prior. 
To alleviate potential biophysical artifacts, such as interpenetration and foot skating, we incorporate a set of physical optimization constraints:
\begin{equation}
    \small
    L = \lambda_\text{joint}L_\text{joint} + \lambda_\text{smooth}L_\text{smooth} + \lambda_\text{pen}L_\text{pen} + \lambda_\text{phy}L_\text{phy},
\end{equation}
where $\lambda$s are weighting factors of each loss function and $L_\text{smooth}$ is a first-order smoothness term:
\begin{equation}
    L_\text{smooth} = \sum_t \Vert \hat{\Theta}_{2:t}-\hat{\Theta}_{1:t-1} \Vert_1 + \sum_t \Vert \hat{\mathbf{K}}^\text{3D}_{2:t}-\hat{\mathbf{K}}^\text{3D}_{1:t-1} \Vert_1,
\end{equation}
where $\hat{\Theta}_i$ and $\hat{\mathbf{K}}^\text{3D}_i$ represent the SMPL-X parameters and joints of the $i$-th frame, respectively. 
To mitigate mesh interpenetration, we utilize a collision penalizer~\cite{collision}, denoted as $L_{\text{pen}}$. 
Additionally, we incorporate the physical loss $L_\text{phy}$ based on PhysCap~\cite{physcap} to prevent implausible poses.


\begin{figure*}
  \begin{center}
      \includegraphics[width=\textwidth]{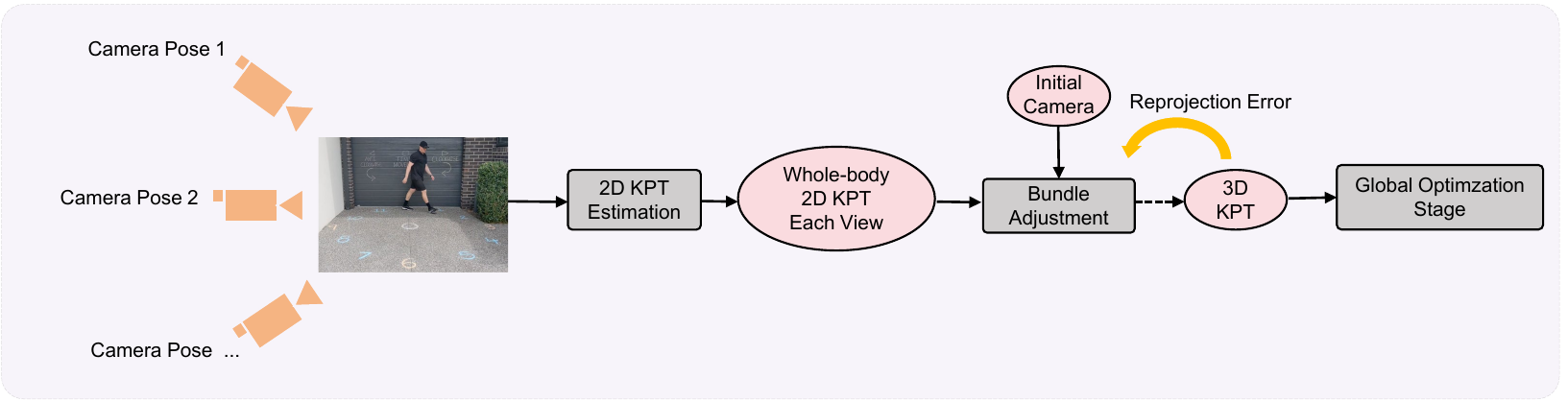}
  \end{center}
  \vspace{-0.2in}
  \caption{Our annotation pipeline can support input from any number of viewpoints. 
  By simultaneously optimizing a fixed set of full-body keypoints and camera poses, our pipeline ensures stable 3D keypoints and motion.}

  \label{fig:mvp}
  \vspace{-0.23in}
\end{figure*}

\noindent\textbf{Extension to Multi-view.} 
For multi-view video inputs, we employ bundle adjustment to calibrate and refine the camera parameters and 3D keypoints. As shown in Fig.~\ref{fig:mvp}, the reprojection error is computed between the estimated 3D and 2D keypoints for each viewpoint according to the camera parameters, jointly optimizing the camera poses and human pose.
To improve stability, we implement temporal smoothing and incorporate 3D bone length constraints during the triangulation process, providing a more robust and accurate estimation of human pose across multiple viewpoints.

\vspace{-0.1in}
\subsection{Global Trajectory Optimization}
%
\noindent\textbf{Camera Trajectory Estimation.} 
To accurately estimate human trajectories in world coordinates, it is essential to obtain precise global camera poses from monocular RGB videos.
%
%
Similar to SLAHMR~\cite{ye2023decoupling}, we utilize DROID-SLAM~\cite{teed2021droid} to estimate the camera pose of the input RGB videos. 
%
For each frame $\textbf{I}_{i}$, DROID-SLAM firstly selects frames containing information exceeding a threshold $\delta_{inf}$ as keyframes $\textbf{I}_{j}$. Then, DROID-SLAM computes 2D revision flows $\textbf{r}_{ij} \in \mathbb{R}$ and their weights $\textbf{w}_{ij}\in \mathbb{R}^{H\times W \times 2}$ relative to keyframes. Given pixel coordinates $\textbf{p}_i \in \mathbb{R}^{H \times W \times 2}$, it calculates the dense correspondence field $\textbf{p}_{ij}$ using:
\begin{equation}
    \textbf{p}_{ij} = \Pi_{c}(G_{ij} \circ \Pi_{c}^{-1}(\textbf{p}_i,\textbf{d}_i)),
\end{equation}
where $\Pi_{c}$ is the camera model projecting 3D points onto the image plane, and $\Pi_{c}^{-1}$ maps pixel coordinates and inverse depth $\textbf{d}_i \in \mathbb{R}^{h \times w}$ to 3D space. The corrected correspondence is $\textbf{p}_{ij} = \textbf{p}_i + \textbf{r}_{ij}$.
%
%
%

DROID-SLAM then optimizes the relative camera pose $G_{ij} \in SE(3)$ and depth $\textbf{d}_i$ using dense bundle adjustment (DBA), minimizing the weighted flow re-projection error:
\begin{equation}\label{eq:droid_obj}
    E(G, d) = \sum_{i, j} ||\textbf{p}_{ij} - \Pi_{c}(G_{ij} \circ \Pi_{c}^{-1}(\textbf{p}_i,\textbf{d}_i)) ||^2_{\Sigma_{ij}}, \Sigma_{ij}=diag\textbf{w}_{ij}
\end{equation}
%
%
%
Finally, global bundle adjustment and loop closure refine the results. The predicted confidence $\textbf{w}_{ij}$ help mitigate errors from uncertain correspondences, improving robustness against minor moving entities.

However, when dynamic objects occupy a large area, the reliability of predicted confidence levels decreases, affecting the accuracy of camera motion estimation. 
TRAM~\cite{wang2024tram} addresses this with a dual mask process, which mask both the input image $\hat{\textbf{I}}_{i}$ = mask($\textbf{I}_{i}$) and the predicted weight $\hat{\textbf{w}}_{ij}$ = mask($\textbf{w}_{ij}$) using SAM~\cite{kirillov2023sam}, where "mask" sets specific regions to zero.
we propose a new mask strategy in DROID-SLAM with two key differences: 1)We use bounding boxes from our annotation pipeline to mask and exclude moving human at the feature extraction level; 2) Besides masking input image and predicted weight, we also mask the pixel coordinates $\hat{\textbf{p}_{i}}$ = mask ($\textbf{p}_{i}$) and 2D revision flow $\hat{\textbf{r}_{ij}}$ = mask ($\textbf{r}_{ij}$) to fully exclude moving human during DBA process.
%
%
%

Nevertheless, masking input images reduces the number of usable pixels for DROID-SLAM, especially when large portions of the frame are occupied by the human object.  
To address this issue, we introduce a scale factor $\lambda_{\text{kf}}$
\begin{equation}
    \vspace{-0.07in}
    \lambda_{\text{kf}} = 1 - \frac{S_{bbox}}{S_{frame}},
\end{equation}
where $S_{bbox}$ and $S_{frame}$ are the areas of the bounding box and frame, respectively.
%
%
To mitigate mask-induced artifacts, we apply a scaling factor $\lambda_{\text{kf}}$ to $\delta_{inf}$ during keyframe extraction, favoring frames with smaller bounding box.

Additionally, we mask pixel coordinates and 2D revision flow as they influence both the calculation of $\textbf{p}_{ij}$ and the extraction of motion features, which is another core element during the calculation of DBA layer. Thus, we have masked correspondence field $\hat{\textbf{p}_{ij}} = \hat{\textbf{p}_{i}} + \hat{\textbf{r}_{ij}}$. 
%
The refined DBA optimization objective is then shown as 
\begin{equation}
    \vspace{-0.05in}
    E(\hat{\textbf{G}}, \hat{\textbf{d}}) = \sum_{i, j} ||\hat{\textbf{p}}_{ij} - \Pi(\hat{\textbf{G}}_{ij} \circ \Pi^{-1}_{c}({\textbf{p}}_i,\hat{\textbf{d}}_i)) ||^2_{\Sigma_{ij}}, \Sigma_{ij}=diag\hat{\textbf{w}}_{ij},
\end{equation}
where $\hat{\textbf{G}}$ and $\hat{\textbf{d}}$ denote the updated camera pose and depth with masked inputs, respectively. 
Thus, we can derive the camera rotation $R$ and camera translation $T$ from $\hat{\textbf{G}}$.\\
\noindent\textbf{Human Trajectory Refinement.} 
To improve the consistency and accuracy of the estimated global trajectory, we adopt a global trajectory optimization strategy similar to SLAHMR~\cite{ye2023decoupling}, using the global motions and camera poses from our masked DROID-SLAM to align with video evidence.
$\textbf{J}_t$ denotes joints of human in world coordinates at frame $t$,
\begin{equation}
    \vspace{-0.07in}
    \textbf{J}_t = M(\Phi_t, \Theta_t, \beta) + \Gamma_t),
\end{equation}
where $\Phi_t$ represents global orientation, and $\Gamma_t$ represents root translation.
We define joint reprojection loss as
\begin{equation}
    \vspace{-0.07in}
    L_{\text{data\_G}} = \Sigma_{t=1}^T \psi_{t}^{i} \rho(\Pi_K(R_t * J_t^{\text{\uppercase\expandafter{\romannumeral1}}} + \alpha T_t)-k_t), 
\end{equation}
and joint smoothness loss as
\begin{equation}
    \vspace{-0.07in}
    L_{\text{smooth\_G}} = \Sigma_{t=1}^T || J_t^{\text{\uppercase\expandafter{\romannumeral1}}} - J_{t+1}^{\text{\uppercase\expandafter{\romannumeral1}}}) ||^2,
\end{equation}
where $k_t$ means the detected 2D keypoint. $\Pi_K([x_1, x_2, x_3]^T) = K[\frac{x_1}{x_3}, \frac{x_2}{x_3}, 1]^T$ is the perspective camera projection with camera intrinsic matrix $K \in \mathbb{R}^{2 \times 3}$, and $\rho$ is the robust Geman-McClure function~\cite{stuart1987statistical}.
As camera scale $\alpha$, shape $\beta$ and SMPL-X parameters $\Theta$ have been derived from the previous stage, we only optimize $\Phi_t$ and $\Gamma_t$ at this stage. 
The final objective function at this stage is,
\begin{equation}
    \vspace{-0.07in}
    \min_{\{\Phi_t, \Gamma_t\}_{t=1}^{T}\}} \lambda_{\text{data\_G}} L_{\text{data\_G}} + \lambda_{\text{smooth\_G}} L_{\text{smooth\_G}}
\end{equation}

For next stage optimization, in line with the approach illustrated in~\cite{reducingskate}, we introduce a ground detector for localizing ground contact events of human feet and use it to impose a physical constraint for optimization of the whole human dynamics in videos. The detector estimates the probability of ground contact $c(j)$ being 0 or 1 for each joint $j$. 
We add a zero velocity constraint for frames where foot joints are in contact with the ground $g$ to avoid unrealistic foot-skating.
For other frames, we will set the velocity in the previous frame.
The loss function is then shown as
\vspace{-0.07in}
\begin{align}
    L_{\text{skate}} = \sum_{t=1}^{T} &\sum_{j=1}^{J} c_{t}(j) \| \mathbf{J}_{t}^{\text{\uppercase\expandafter{\romannumeral2}}}(j) - \mathbf{J}_{t+1}^{\text{\uppercase\expandafter{\romannumeral2}}}(j) \| + \notag \\
    & (1-c_{t}(j)) \| \mathbf{J}_{t}^{\text{\uppercase\expandafter{\romannumeral1}}}(j) - \mathbf{J}_{t+1}^{\text{\uppercase\expandafter{\romannumeral1}}}(j) \|.
\end{align}
where \uppercase\expandafter{\romannumeral1}, \uppercase\expandafter{\romannumeral2} illustrate the smoothing stage and foot contact optimization stage.
Meanwhile, we ensure the distance between joints and ground remains below a threshold $\delta$,
\vspace{-0.07in}
\begin{equation}
    L_{\text{con}} = \sum_{t=1}^{T} \sum_{j=1}^{J} c_{t}(j) \max(d(\mathbf{J}_{t}^{\text{\uppercase\expandafter{\romannumeral2}}}(j), g) - \delta, 0)
\end{equation}
where $d(p, g)$ defines the distance between the point $p \in \mathbb{R}^{3}$ and the plane $g \in \mathbb{R}^{3}$.
$g$ is optimized during all timesteps.
Therefore, our loss function at this stage can be denoted as,
\vspace{-0.07in}
\begin{equation}\label{eq:stage3loss}
    \min_{\{\Phi_t, \Gamma_t\}_{t=1}^{T}\}} \lambda_{\text{data\_G}} L_{\text{data\_G}} + \lambda_{\text{skate}} L_{\text{skate}} + \lambda_{\text{con}} L_{\text{con}}.
\end{equation}
The optimization process of Eq.~\eqref{eq:stage3loss} is similar to SLAHMR.
%


\noindent\textbf{Human Verification.} 
To ensure quality, we manually checked the annotation by removing the motions that either deviated from the video evidence or displayed apparent biophysical inconsistencies incompatible with natural human movement.

\subsection{Obtaining Whole-body Motion Descriptions}
\label{sec:lable_text}

\noindent\textbf{Sequence motion labels.} 
The videos in \newdataname were collected from online sources and existing datasets. 
For action-related datasets~\cite{cai2022humman,chung2021haa500,li2021aist,liu2019ntu,amass,taheri2020grab}, we use the action labels as one of the sequence semantic labels. Additionally, feeding the videos into GPT-4V~\cite{openai2024gpt4v} and filtering the human action descriptions as supplemental texts. 
When videos contain semantic subtitles, EasyOCR automatically extracts semantic information. For online videos, we also use the search queries generated from~\cite{jiao2023chatgpt} as semantic labels.
Videos without available semantic information, such as EgoBody~\cite{zhang2022egobody}, are manually labeled using the VGG Image Annotator (VIA)~\cite{dutta2019vgg}.
For face database BAUM~\cite{baum}, we use the facial expression labels provided by the original creator.
%

\begin{figure*}[h]
\begin{center}
    \includegraphics[width=0.97\textwidth]{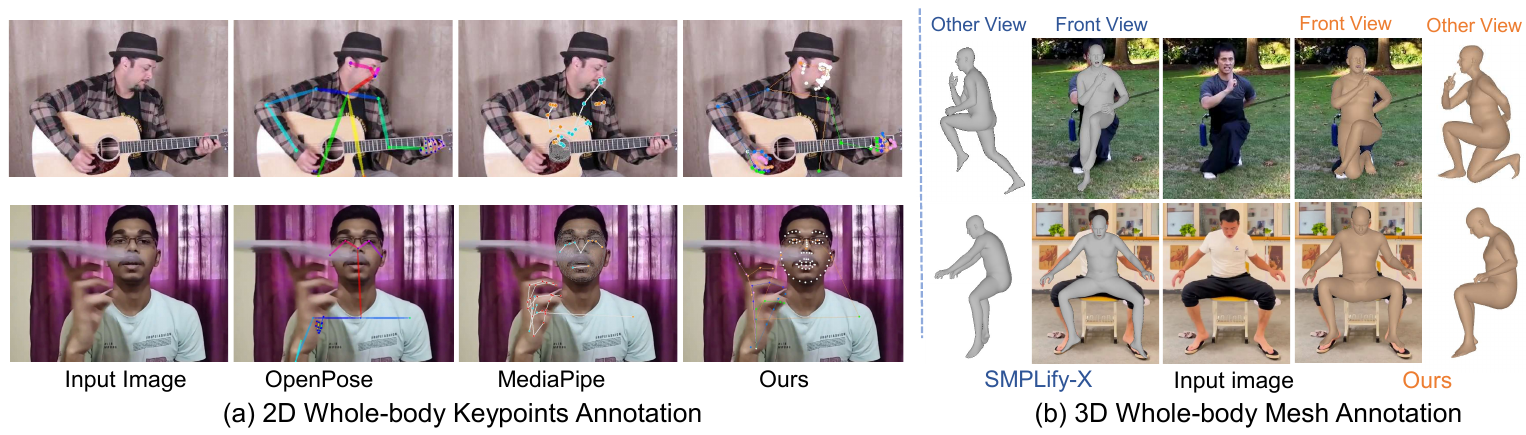}
\end{center}
\vspace{-0.2in}
\caption{\small Qualitative comparisons of (a) 2D keypoints annotation with widely used methods~\cite{openpose, mediapipe} and (b) the 3D mesh annotation with the popular fitting method~\cite{smpl-x} with ours.}
\label{figure:motion_anno_compare}
\vspace{-0.3in}
\end{figure*}

\noindent\textbf{Whole-body pose descriptions.} The generation of fine-grained pose descriptions for each pose involves three distinct parts: face, body, and hand, as shown in Fig.~\ref{figure:motion_anno_compare}(a).
\emph{Facial expression labeling} uses the emotion recognition model EMOCA~\cite{emoca} pre-trained on AffectNet~\cite{mollahosseini2017affectnet} to classify the emotion. 
\emph{Body-specific descriptions} utilizes the captioning process from PoseScript~\cite{posescript}, which generates synthetic low-level descriptions in natural language based on given 3D keypoints. The unit of this information is called pose codes, such as \textit{`the knees are completely bent'}. A set of generic rules based on fine-grained categorical relations of the different body parts are used to select and aggregate the low-level pose information.  The aggregated pose codes are then used to produce textual descriptions in natural language using linguistic aggregation principles.
\emph{Hand gesture descriptions} extends the pre-defined posecodes from body parts to fine-grained hand gestures. 
We define six elementary finger poses via finger curvature degrees and distances between fingers to generate descriptions, such as \textit{`bent'} and \textit{`spread apart'}.
We calculate the angle of each finger joint based on the 3D hand keypoints and determine the corresponding margins.
For instance, if the angle between $\vec{\mathbf{V}} (\mathbf{K}_\text{wrist}, \mathbf{K}_\text{fingertip}) $ and $\vec{\mathbf{V}} (\mathbf{K}_\text{fingertip}, \mathbf{K}_\text{fingeroot}) $ falls between 120 and 160 degrees, the finger posture is labeled as \textit{`slightly bent'}.
We show an example of the annotated text labels in Fig.~\ref{figure:motion_anno_compare}(b). 

\noindent\textbf{Summary.} Based on the above annotations, we build \newdataname, which has 120.5K clips with 19.5M SMPL-X poses and the corresponding pose and semantic text labels.

\section{Experiment}
In this section, we validate the performance of our motion annotation pipeline and the effectiveness of \newdataname in several downstream tasks.
%

\begin{figure}[h]
  \vspace{0.08in}
  \centering
    \subfloat[\small Evaluation result on the COCO-Wholebody~\cite{coco_wholebody} dataset.]{
      \scalebox{0.81}{
      \begin{tabular}{l|cc|cc|cc}
        \toprule
        \rowcolor{lightgray}
        & \multicolumn{2}{c|}{hand $\uparrow$} 
        & \multicolumn{2}{c|}{face $\uparrow$} 
        & \multicolumn{2}{c}{whole-body $\uparrow$} \\
        \rowcolor{lightgray}
        \multirow{-2}{*}{{Method}} &
        {AP} & {AR} & {AP} & {AR} & {AP} & {AR} \\
        \midrule
        OpenPose~\cite{openpose} & 38.6 & 43.3 & 76.5 & 84.0 & 44.2 & 52.3  \\
        HRNet~\cite{hrnet}  & 50.3 & 60.2 & 73.7 & 80.9 & 58.2 & 67.1  \\
        ViTPose~\cite{YufeiXu2022ViTPoseSV}  & 47.4 & 59.4 & 59.8 & 70.7 & 57.7 & 69.4  \\
        RTMPose-~L\cite{rtmpose}  & 52.3 & 60.0 & 84.4 & 87.6 & 63.2 & 69.4  \\
        Ours & \textbf{64.9} & \textbf{74.0} & \textbf{91.6} & \textbf{94.4} 
        & \textbf{73.5} {\color{red}$\uparrow_{16.3\%}$} & \textbf{80.3} {\color{red}$\uparrow_{15.7\%}$} \\
        \bottomrule
        \end{tabular}}
      }\hfill
    \vspace{-0.05in}
    \subfloat[\small Reconstruction error on the EHF~\cite{smpl-x} dataset.]{
      \scalebox{1.10}{
        \scriptsize
        \begin{tabular}{lccc}
          \toprule
          \rowcolor{lightgray}
          Method & PA-MPJPE $\downarrow$ & PA-MPVPE $\downarrow$ & MPVPE $\downarrow$  \\
          \midrule
          Hand4Whole~\cite{GyeongsikMoon2020hand4whole} & 58.9 & 50.3 & 79.2 \\
          OSX~\cite{osx} & 55.6& 48.7 & 70.8 \\
          PyMAF-X~\cite{pymafx} & 52.8 & 50.2 & 64.9 \\
          SMPLify-X~\cite{smpl-x} & 62.6 & 52.9 & - \\
          \midrule
          Ours & 33.5 & 31.8 & 44.7{\color{red}$\downarrow_{30.1\%}$} \\
          Ours w/GT 3Dkpt & \textbf{23.9} & \textbf{19.7} &  \textbf{30.7} {\color{red}$\downarrow_{52.7\%}$}\\
          \bottomrule
          \end{tabular}}   
    }
  \vspace{-0.1in}
  \captionof{table}{\small Evaluation of motion annotation pipeline on (a) 2D keypoints and (b) 3D SMPL-X datasets.}
  \label{tab:annot_eval}
  \vspace{-0.25in}
\end{figure}
\subsection{Evaluation of the Motion Annotation Pipeline}
\label{sec:exp_eval}
In this part, we first validate the accuracy of our
motion annotation pipeline on the 2D keypoints and 3D SMPL-X datasets.
Secondly, we evaluate the precision of the estimated camera and human trajectories using our pipeline and compare it with other methods.
\subsubsection{Evaluation of Shot Detection}
\begin{figure}[H]
  \begin{center}
      \includegraphics[width=0.48\textwidth]{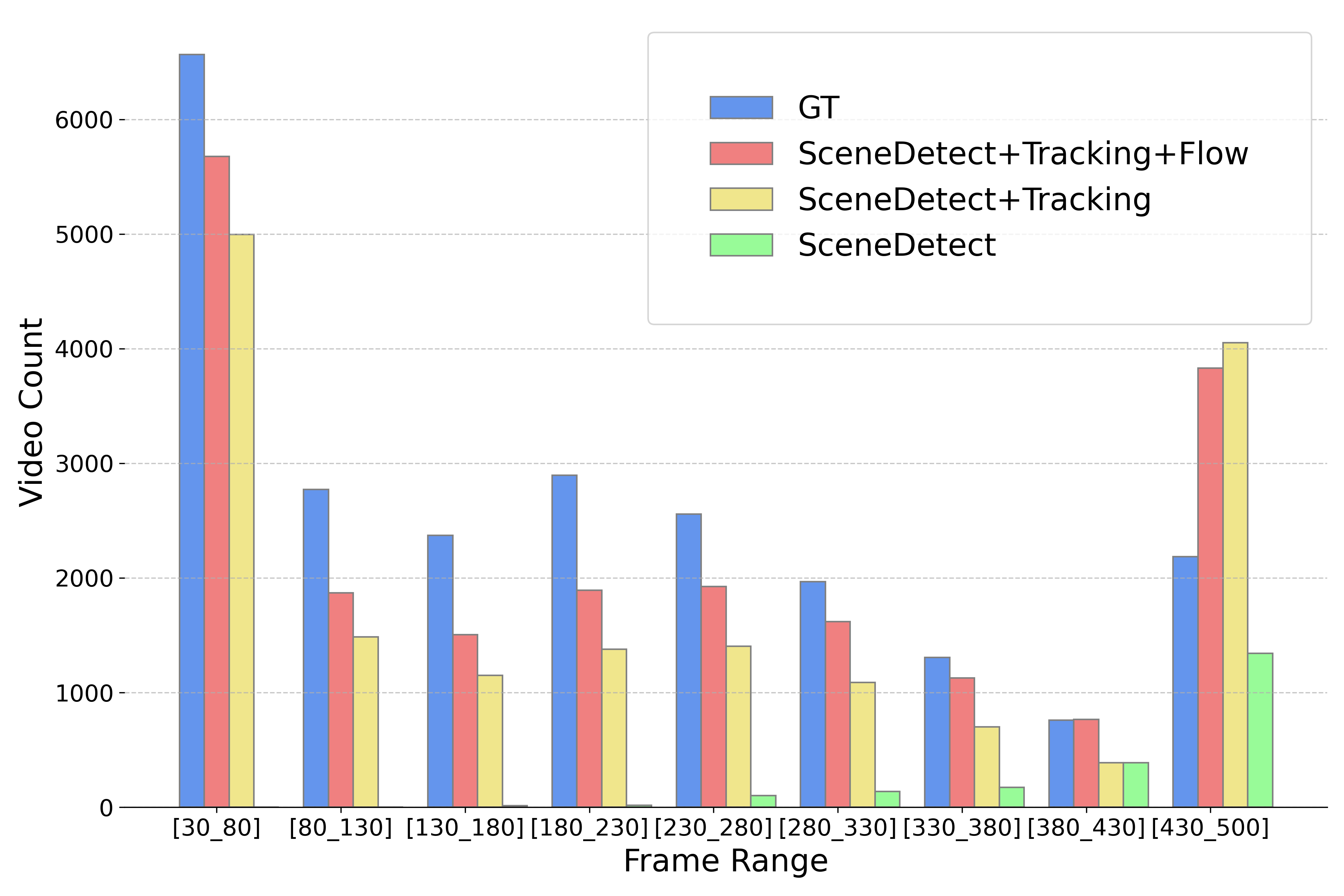}
  \end{center}
  \vspace{-0.22in}
  \caption{\small Frame distribution across the entire \textit{Game Motion} Dataset using various clip method combinations. [30\_80] means the frame length ranges from 30 to 80. The combined shot detection method (SceneDetect+Tracking+optical Flow) closely matches the ground truth (GT) distribution.}
  \label{fig:framedistribution}
\end{figure}

We evaluate the performance of Shot Detection algorithm in this section. Fig.~\ref{fig:framedistribution} shows the frame distribution after applying various shot detection algorithms to the original \textit{Game Motion} Dataset. It reveals that the combined algorithm, incorporating scene detection, tracking, and optical flow perception, most accurately identifies shot boundaries. 
This approach closely matches the distribution obtained through manual inspection, providing more robust detection than scene detection alone.

\subsubsection{Evaluation of Whole-body Pose Estimation}
\noindent \textbf{2D Keypoints Annotation.} We evaluate the proposed 2D keypoint annotation method on the COCO-WholeBody~\cite{coco_wholebody} dataset and compare the evaluation result with four SOTA keypoints estimation methods ~\cite{openpose, hrnet, YufeiXu2022ViTPoseSV, rtmpose}.
We use the same input image size of $256\times 192$ for all methods to ensure a fair comparison. 
From Tab.~\ref{tab:annot_eval}(a), our annotation pipeline significantly surpasses existing methods by over 15\% average precision. Additionally, we provide qualitative comparisons in Fig.~\ref{figure:motion_anno_compare}(a), illustrating the robust performance of our method, especially in challenging and occluded scenarios. 

\noindent \textbf{3D SMPL-X Annotation.}
We evaluate our learning-based fitting method on the EHF~\cite{smpl-x} dataset and compare it with four open-sourced human mesh recovery methods. Following previous works,
we employ mean per-vertex error (MPVPE), Procrusters-aligned mean per-vertex error (PA-MPVPE), and Procrusters-aligned mean per-joint error (PA-MPJPE) as evaluation metrics (in mm).
Results in Tab.~\ref{tab:annot_eval}(b) demonstrate the superiority of our progressive fitting method (over 30\% error reduction). 
Specifically, PA-MPVPE is only 19.71 mm when using ground-truth 3D keypoints as supervision. 
Fig.~\ref{figure:motion_anno_compare}(b) shows the annotated mesh from front and side view, indicating reliable 3D SMPL-X annotations with reduced depth ambiguity. More results are presented in Appendix due to page limits.
\subsubsection{Evaluation of the Global Trajectory Optimization}
\noindent \textbf{Overview.} This section aims to demonstrate the efficacy of the proposed mask operation on DROID-SLAM and global stage optimizing strategy from three perspectives.
Firstly, we assess its performance in global camera trajectory estimation on videos shot by static cameras, which constitute over 90\% of the \newdataname dataset.
Secondly, we evaluate its effectiveness in global camera trajectory estimation on videos featuring moving cameras.
Lastly, within the context of moving camera scenarios, we estimate global human trajectories and compare them with other methods.
\begin{figure}[H]
    \vspace{-0.15in}
    \centering
    \includegraphics[width=\linewidth]{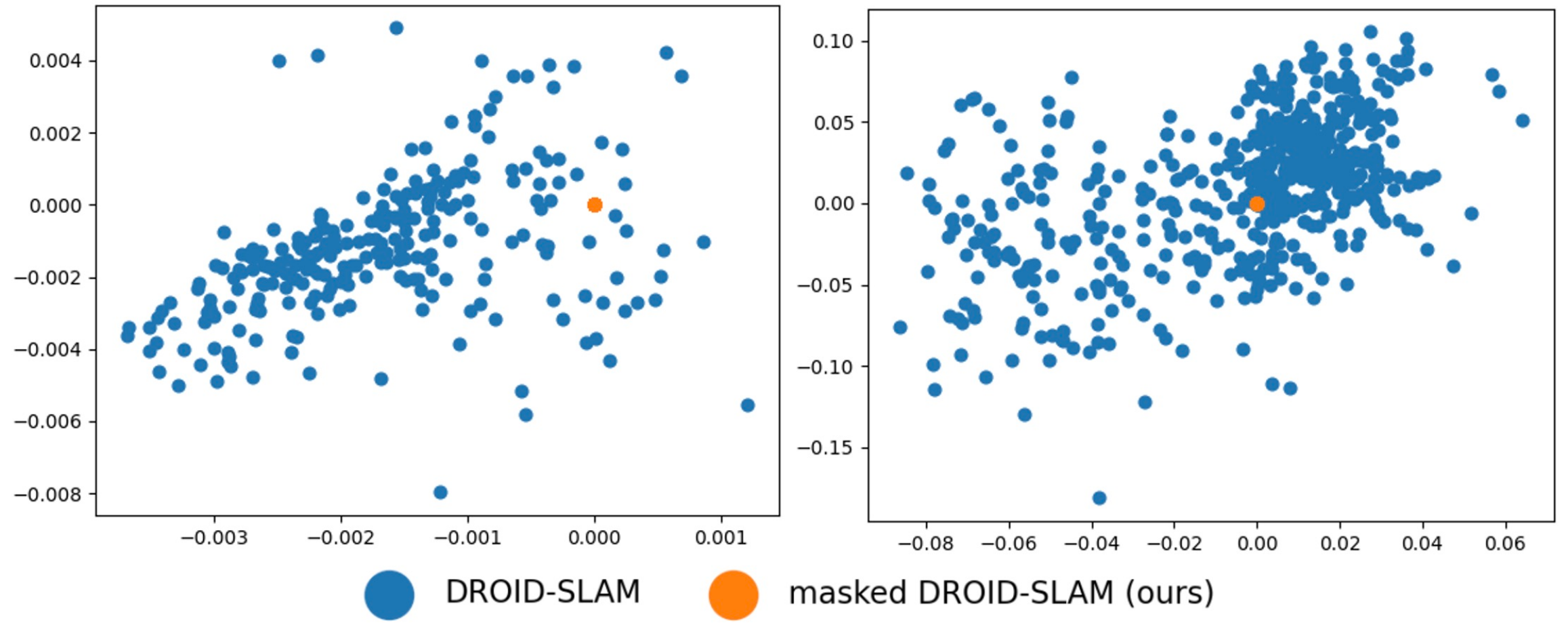}
    \vspace{-0.27in}
    \caption{\small Camera trajectory comparison between original Droid-SLAM and our method on videos with static camera pose. 
    Unlike the original DROID-SLAM, whose output dots (blue) indicate it as a moving camera, the masked DROID-SLAM (orange) is represented by a single point and is perceived as static.}
    \label{fig:trans_compare}
    \vspace{-0.13in}
\end{figure}
\noindent \textbf{Static Camera.} For this evaluation, we select a subset from the \textit{Dance} dataset within the \newdataname dataset. 
For illustration, two videos with static camera scenes are depicted in Fig.~\ref{fig:trans_compare}. 
It suggests that our mask operation effectively eliminates the influence of moving objects, thereby demonstrating the efficacy and accuracy of our pipeline in annotating videos with static cameras.
%
\begin{figure}[H]
    \vspace{-0.15in}
    \centering
    \includegraphics[width=\linewidth]{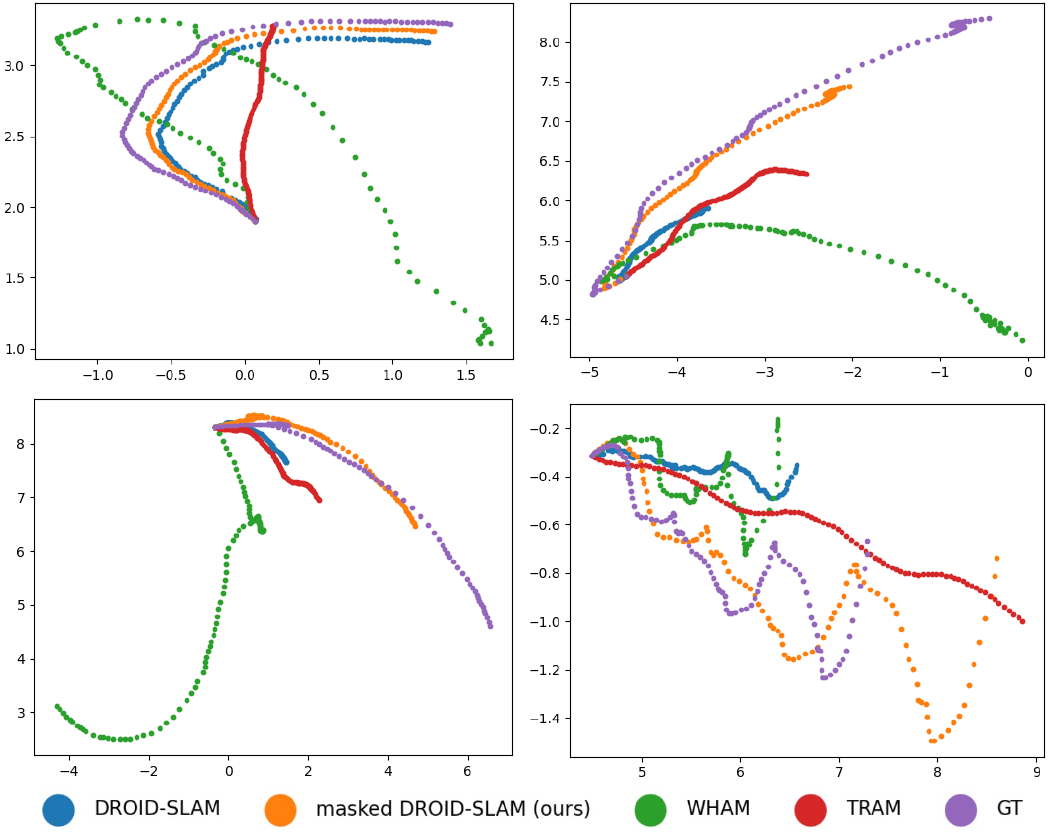}
    \vspace{-0.27in}
    \caption{\small Global Camera Trajectory Comparison on EMDB: Our method yields more accurate global camera trajectories compared to DROID-SLAM, WHAM, and TRAM.}
    \label{fig:cam_trans}
\end{figure}
\begin{figure}[H]
    \vspace{-0.26in}
    \centering
    \includegraphics[width=\linewidth]{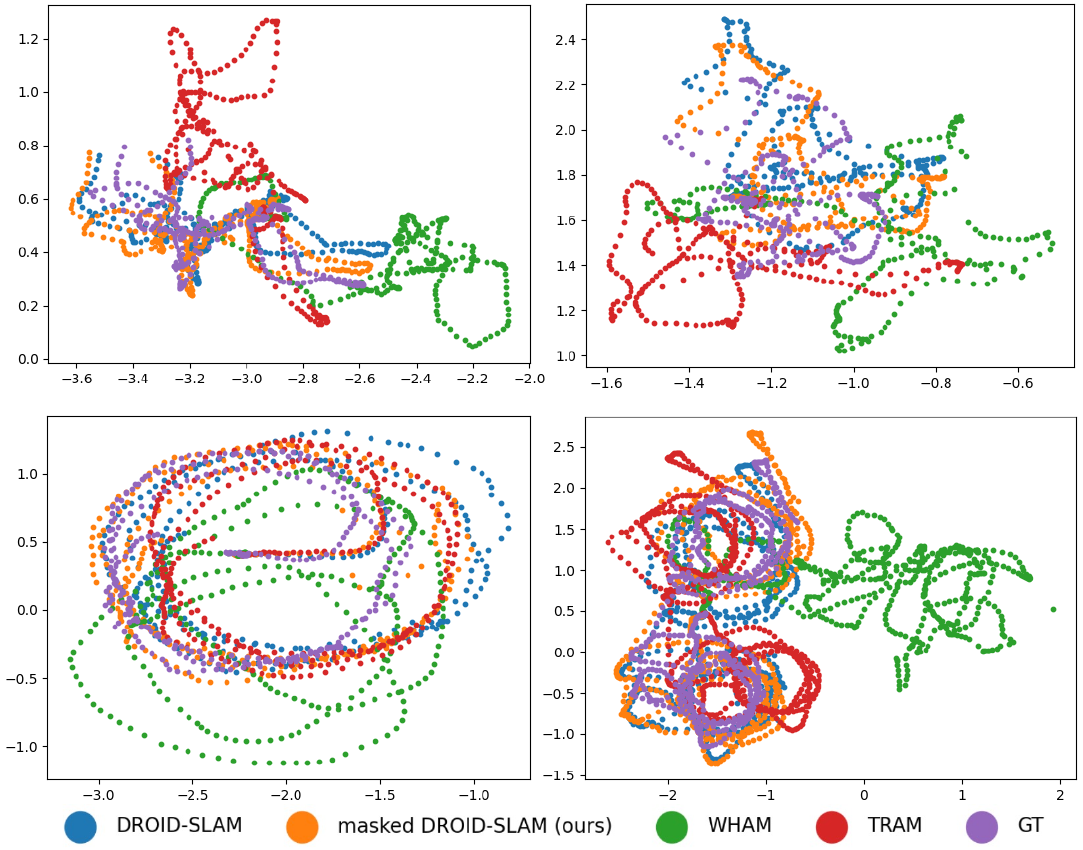}
    \vspace{-0.27in}
    \caption{\small Global human trajectory comparison on EMDB. Compared to DROID-SLAM, our precise camera trajectory estimation results in finer global human trajectory predictions, as both use the same method for global human trajectory estimation. Our method is also more similar as GT than WHAM and TRAM.}
    \label{fig:human_trans}
    \vspace{-0.17in}
\end{figure}
\noindent \textbf{Moving Camera.} For the evaluation of moving cameras, we utilize the EMDB dataset~\cite{kaufmann2023emdb}, which provides ground truth (GT) camera and human trajectories.
We conduct experiments using DROID-SLAM, WHAM~\cite{shin2024wham} (which employs DPVO~\cite{teed2023deep} for camera trajectory estimation), TRAM (which applies their proposed mask operation to DROID-SLAM), and our method (masked DROID-SLAM) on the EMDB dataset.
Selected comparisons of the trajectories are presented in Fig.~\ref{fig:cam_trans}.
The results demonstrate that our method achieves a more accurate scale and direction than the original DROID-SLAM and WHAM. 
Furthermore, compared to TRAM, our method produces more similar trajectories as GT.
\begin{figure}[htbp]
  \vspace{-0.1in}
  \centering
    \scalebox{1.5}{
        \tiny
        \begin{tabular}{lcc}
          \toprule
          \rowcolor{lightgray}
          Method\footnotemark[1] & ATE (Cam Traj) $\downarrow$ & RTE (Human Traj) $\downarrow$ \\
          \midrule
          DROID-SLAM~\cite{ye2023decoupling} & 0.217 & 4.773 \\
          WHAM~\cite{shin2024wham} & 0.305 & 3.917 \\
          TRAM~\cite{wang2024tram} & 0.314 & \textbf{2.943} \\
          \midrule
          Ours& \textbf{0.202} &  3.006 \\
          \bottomrule
        \end{tabular}}
      \vspace{-0.07in}
      \captionof{table}{\small Evaluation of camera and human trajectory on the EMDB dataset. ATE [m] and RTE [m] are used for the evaluation of global camera and human trajectory, respectively.
      }
    \label{tab:trans_eval}
    \vspace{-0.1in}
\end{figure}

\noindent \textbf{Global Human Trajectory Estimation.} We also utilize the EMDB dataset to evaluate the performance of our method and other approaches in estimating global human trajectories.
As depicted in Fig.~\ref{fig:human_trans}, our method produces a trajectory shape that closely matches the GT when compared to WHAM and TRAM.
The primary distinction between DROID-SLAM and our method is the mask operation, which results in more accurate global human trajectories in both scale and shape.
This comparison underscores the effectiveness of our proposed mask operation, ultimately leading to a more accurate estimation of global human trajectories.
To precisely evaluate the performance of our proposed mask operation,we compare the Root Trajectory Error (RTE) and Absolute Trajectory Error (ATE)—both commonly employed for evaluating global camera and human trajectories—across DROID-SLAM, WHAM, TRAM, and our method based on ground truth (GT).
The results are presented in Tab.~\ref{tab:trans_eval}.
These values represent the average deviation of each method from the GT for each frame in the EMDB dataset.
\footnotetext[1]{The performance of WHAM and TRAM listed here are run by using their provided code from Github. 
We use 64 sequences for ATE evaluation and 60 for RTE evaluation from the EMDB dataset.}
%
\subsection{Validation of the Downstream Tasks}
In this section, we validate the effectiveness of \newdataname in whole-body text- and music-driven motion generation, human mesh recovery and pose estimation.
%
\subsubsection{Impact on Text-driven Whole-body Motion Generation}
\label{sec:t2m_exp}
\noindent\textbf{Experiment Setup.} We randomly split \newdataname into train ($80\%$), val ($5\%$), and test ($15\%$) sets. SMPL-X is adopted as motion representation for expressive motion generation.

\noindent\textbf{Evaluation metrics.}
We adopt the same evaluation metrics as \cite{humanml3d}, including Frechet Inception Distance (FID), Multimodality, Diversity, R-Precision, and Multimodal Distance. 
Due to the page limit, we leave more details about experimental setups and evaluation metrics in the appendix.

\begin{table*}[t]
        \centering
	\resizebox{0.95\textwidth}{!}{%
            \scriptsize
		\begin{tabular}{lccccccc}
			\toprule
      \rowcolor{color3}
			 & \multicolumn{3}{c}{R Precision $\uparrow$}& & & & 
       \\ 
      \rowcolor{color3}
			\multirow{-2}{*}{Methods} & \multicolumn{1}{c}{Top 1} & 
      \multicolumn{1}{c}{Top 2} & 
      \multicolumn{1}{c}{Top 3} & 
      \multirow{-2}{*}{FID$\downarrow$} & 
      \multirow{-2}{*}{MM Dist$\downarrow$} & 
      \multirow{-2}{*}{Diversity$\rightarrow$} & 
      \multirow{-2}{*}{MModality}
      \\ 
      \midrule
			Real &
			$0.573^{\pm.005}$ &
			$0.765^{\pm.003}$ &
			$0.850^{\pm.005}$ &
			$0.001^{\pm.001}$&
			$2.476^{\pm.002}$&
			$13.174^{\pm.227}$ &
			- \\
			\midrule
			MDM \cite{Tevet2022HumanMD}&
                $0.290^{\pm.011}$ &
                $0.459^{\pm.010}$ &
                $0.577^{\pm.008}$ &
                $2.094^{\pm.230}$ &
                $6.221^{\pm.115}$ &
                $11.895^{\pm.354}$ &
                $\boldsymbol{2.624}^{\pm.083}$
                \\
      MLD \cite{mld}&
			$0.440^{\pm.002}$ &
			$0.624^{\pm.004}$ &
			$0.733^{\pm.003}$ &
			$0.914^{\pm.056}$ &
			$3.407^{\pm.020}$ &
			$13.001^{\pm.245}$ &
			$2.558^{\pm.084}$
                \\ 
                
      T2M-GPT \cite{t2m-gpt}&
      $0.502^{\pm.004}$ &
      $0.697^{\pm.005}$ &
      $0.791^{\pm.007}$ &
      $0.699^{\pm.012}$ &
      $3.192^{\pm.035}$ &
      $\boldsymbol{13.132}^{\pm.127}$ &
      $2.510^{\pm.027}$ \\ 
      
        MotionDiffuse \cite{motiondiffuse} &
        $\boldsymbol{0.559}^{\pm.001}$ &
        $\boldsymbol{0.748}^{\pm.004}$ &
        $\boldsymbol{0.842}^{\pm.003}$ &
        $\boldsymbol{0.457}^{\pm.007}$ &
        $\boldsymbol{2.542}^{\pm.018}$ &
        $13.576^{\pm.161}$ &
        ${1.620}^{\pm.152}$ \\

			  \bottomrule
		\end{tabular}%
	}
	\vspace{-0.1in}
	\caption{\small Benchmark of text-driven motion generation on \newdataname test set. 
  `$\rightarrow$' means results are better if the metric is closer to the real motions and $\pm$ indicates the $95\%$ confidence interval.} 
	\label{tab:benchmark}
    \vspace{-0.05in}
\end{table*}

\begin{table*}[t]
    \centering
    \resizebox{1.0\textwidth}{!}{
    \begin{tabular}{l|cccc|cccc}
      \toprule
      \rowcolor{lightgray}
      & 
      \multicolumn{4}{c|}{HumanML3D (Test)} & 
      \multicolumn{4}{c}{Motion-X++ (Test)} \\
      \rowcolor{lightgray}
      \multicolumn{1}{c|}{\multirow{-2}{*}{{Train Set}}} &
      {R-Precision$\uparrow$} & 
      {FID$\downarrow$} & 
      {Diversity$\rightarrow$} & 
      {MModality}& 
      {R-Precision$\uparrow$} & 
      {FID$\downarrow$} & 
      {Diversity$\rightarrow$} & 
      {MModality} \\
      \midrule
      Real (GT) & $0.749^{\pm.002}$ & $0.002^{\pm.001}$ & $9.837^{\pm.084}$ & - & $0.850^{\pm.005}$ & $0.001^{\pm.001}$ &  $13.174^{\pm.227}$ & -  \\
      \midrule
      HumanML3D & $0.657^{\pm.004}$ & $1.579^{\pm.050}$ & $10.098^{\pm.052}$ & $2.701^{\pm.143}$ & $0.570^{\pm..003}$ & $12.309^{\pm.127}$ & $9.529^{\pm.165}$ & $\boldsymbol{2.960}^{\pm.066}$ \\
      Motion-X++ &  $\mathbf{0.695}^{\pm.005}$ & $\mathbf{0.999}^{\pm.042}$ & $\mathbf{9.871}^{\pm.099}$ & $\mathbf{2.827}^{\pm.138}$ & $\mathbf{0.733}^{\pm.003}$ & $\mathbf{0.914}^{\pm.056}$ & $\mathbf{13.001}^{\pm.245}$ & $2.558^{\pm.084}$ \\ 
      \bottomrule
      \end{tabular}}
      \vspace{-0.1in}
      \caption{\small Cross-dataset comparisons of HumanML3D and \newdataname. We train MLD on the training set of HumanML3D and \newdataname, respectively, and then evaluate it on their test sets.}
      \label{tab:compare_humanml}
      \vspace{-0.2in}
\end{table*}

\noindent\textbf{Benchmarking Motion-X++.} We train and evaluate four diffusion-based motion generation methods, including MDM~\cite{Tevet2022HumanMD}, MLD~\cite{mld}, MotionDiffuse~\cite{motiondiffuse} and T2M-GPT~\cite{t2m-gpt} on our dataset. Since previous datasets only have sequence-level motion descriptions, we keep similar settings for minimal model adaptation and take semantic label as text input. The evaluation is conducted with 20 runs (except for multimodality with 5 runs) under a $95\%$ confidence interval. From Tab.~\ref{tab:benchmark}, MotionDiffuse demonstrates a superior performance across most metrics. However, it scores the lowest in multimodality, indicating it generates less varied motion. Notably, T2M-GPT achieves comparable performance on our dataset while maintaining high diversity, meaning our large-scale dataset's promising prospects to enhance the GPT-based method's efficacy. MDM gets the highest multimodality score with the lowest precision, indicating the generation of noisy and jittery motions. The highest Top-1 precision is 55.9\%, showing the challenges of \newdataname.
MLD adopts the latent space design, making it fast while maintaining competent results. Therefore, we use MLD to conduct the following experiments to compare \newdataname with the existing largest motion dataset HumanML3D and ablation studies.

\begin{figure}[htbp]
  \centering
  \vspace{-0.15in}
  \includegraphics[width=0.49\textwidth]{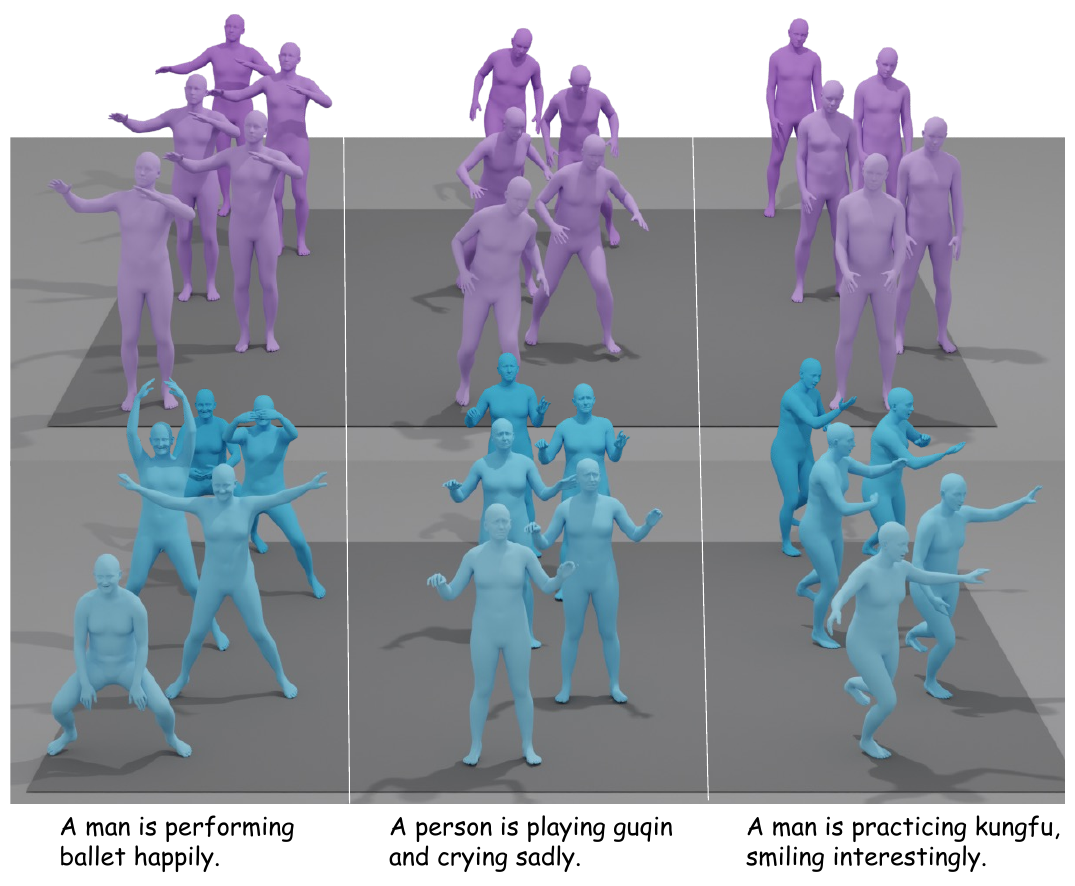}
  \vspace{-0.3in}
  \caption{\small Visual comparisons of motions generated by MLD~\cite{mld} trained on HumanML3D (in purple) or \newdataname (in blue). Please zoom in for a detailed comparison. The model trained with \newdataname can generate more accurate and semantic-corresponded motions.}
  \label{fig:motion_gen}
  \vspace{-0.17in}
\end{figure}

\noindent \textbf{Comparison with HumanML3D.} To validate the richness, expressiveness, and effectiveness of our dataset, we conduct a comparative analysis between \newdataname and HumanML3D, which is the largest existing dataset with text-motion labels. We replace the original vector-format poses of HumanML3D with the corresponding SMPL-X parameters from AMASS~\cite{amass}, and randomly extract facial expressions from BAUM~\cite{baum} to fill in the face parameters. We train MLD separately on the training sets of \newdataname and HumanML3D, then evaluate both models on the two test sets. The results in Tab.~\ref{tab:compare_humanml} reveal some valuable insights. Firstly, \newdataname exhibits greater diversity (\textbf{13.174}) than HumanML3D (\textbf{9.837}), as evidenced by the real (GT) row. This indicates a wider range of motion types captured by \newdataname. Secondly, the model pre-trained on \newdataname and then fine-tuned on the HumanML3D subset performs well on the HumanML3D test set, even better than the intra-data training. These superior performances stem from the fact that \newdataname encompasses diverse motion types from massive outdoor and indoor scenes.  For a more intuitive comparison, we provide the visual results of the generated motion in Fig.~\ref{fig:motion_gen}, where we can clearly see that the model trained on \newdataname excels at synthesizing semantically corresponding motions given text inputs.
These results prove the significant advantages of \newdataname in enhancing expressive, diverse, and natural motion generation. 

\noindent\textbf{Ablation study of text labels.}
In addition to sequence-level semantic labels, the text labels in \newdataname also include frame-level pose descriptions, which is an important characteristic of our dataset. To assess the effectiveness of pose description, we conducted an ablation study on the text labels. The baseline model solely utilizes the semantic label as the text input. Since there is no method to use these labels, we simply sample a single sentence from the pose descriptions randomly, concatenate it with the semantic label, and feed the combined input into the CLIP text encoder. Interestingly, from Tab.~\ref{tab:ablation}, adding additional face and body pose texts brings consistent improvements, 
and combining whole-body pose descriptions results in a noteworthy $38\%$ reduction in FID.
These results validate that the proposed whole-body pose description contributes to generating more accurate and realistic human motions. More effective methods to utilize these labels can be explored in the future.
\begin{figure}[t]
  \begin{minipage}{0.5\textwidth}
  \centering
  \resizebox{\linewidth}{!}{
      \makeatletter\def\@captype{table}\makeatother
      \scriptsize
      \begin{tabular}{c c c c c}
        \toprule
        \rowcolor{lightgray}
        Semantic & \multicolumn{3}{c}{Pose Description} 
        &  \\
        \rowcolor{lightgray}
        Label & face text & body text & hand text  & 
        \multirow{-2}{*}{~FID$\downarrow$~} \\
        \midrule
        \checkmark &  &  &   & $0.914^{\pm.056}$ \\
        \checkmark &\checkmark  &  &  & $0.784^{\pm.032}$ \\
        \checkmark  & \checkmark & \checkmark &  & $0.671^{\pm.016}$ \\
        \checkmark  &\checkmark  & \checkmark & \checkmark & $\mathbf{0.565}^{\pm.036}$ \\
        \bottomrule
        \end{tabular}
      }
      \vspace{-0.1in}
      \captionof{table}{\small Ablation study of text inputs.}
      \vspace{0.08in}
      \label{tab:ablation}
  \end{minipage}
  \hfill
  \begin{minipage}{0.5\textwidth}
    \centering
    \resizebox{\linewidth}{!}{
        \makeatletter\def\@captype{table}\makeatother
        \footnotesize
        \begin{tabular}{l|ccc|ccc}
          \toprule
          \rowcolor{lightgray}
          & \multicolumn{3}{c|}{\boldmath{}{EHF~\cite{smpl-x} $\downarrow$}\unboldmath{}} & \multicolumn{3}{c}{\boldmath{}{AGORA~\cite{action2video} $\downarrow$}\unboldmath{}} \\
          \rowcolor{lightgray}
          \multicolumn{1}{c|}{\multirow{-2}{*}{{Method}}} & {all} & {hand} & {face} & {all} & {hand} & {face} \\
          \midrule
          w/o Motion-X++ & 79.2 & 43.2 &25.0 &185.6 & 73.7 & 82.0   \\
          w/ Motion-X++ & \textbf{73.0} & \textbf{41.0} & \textbf{22.6} &\textbf{184.1} & \textbf{73.3} & \textbf{81.4}  \\
          \bottomrule
          \end{tabular}
        }
        \vspace{-0.1in}
        \captionof{table}{\small Mesh recovery errors of Hand4Whole~\cite{GyeongsikMoon2020hand4whole}
        using different training datasets. MPVPE (mm) is reported.}
        \label{tab:mesh_recovery}
    \end{minipage}
    \vspace{-0.25in}
\end{figure}
%
\subsubsection{Impact on Music-driven Motion Generation}
\noindent \textbf{Experiment Setup.} Our annotated Dance and AIST datasets contain diverse, natural dance motion sequences and corresponding audio. We train and evaluate FineDance~\cite{finedance} and EDGE~\cite{edge} with our annotated AIST and part of FineDance data. \newdataname randomly separated into the train ($90\%$), validation ($5\%$), and test ($5\%$) sets. SMPL-H is adopted as the motion representation for training and testing.

\noindent \textbf{Evaluation metrics.}
We employ the widely used music-to-dance metrics to evaluate different methods on our dataset.
We use \textit{Diversity}, \textit{Beat Aligned}, and \textit{Physical Foot Contact score~\cite{edge}}(PFC) to evaluate the average feature distance between generated dances for different music inputs, the alignment of generated dances with the music beat, physically-inspired metric targeting foot sliding, respectively.

\noindent \textbf{Evaluation results.}
The results in Tab.~\ref{tab:music2dance} indicate that models trained with \newdataname data generate more diverse and natural motions, and models trained using FineDance method achieve \textit{PFC} metric that closer to GT, trained with our annotated AIST and~\cite{finedance} demonstrate strong beat alignment, even with in-the-wild audio. Visualization results are provided in Appendix Fig.~\ref{figure:dance}, indicating less foot sliding and more physically plausible motions.
\subsubsection{Impact on Whole-body Human Mesh Recovery}
\label{sec:mesh_recovery}
As discovered in this benchmark~\cite{pang2022benchmarking}, the performance of mesh recovery methods can be significantly improved by utilizing high-quality pseudo-SMPL labels. \newdataname provides a large volume of RGB images and well-annotated SMPL-X labels.
%
\begin{table}[htbp]
  \vspace{-0.1in}
  \centering
    \scalebox{1.6}{
    \tiny
    \begin{tabular}{l|c|c|c}
     \toprule
     \rowcolor{lightgray} 
    Method    & PFC $\downarrow$ & Beat-Align $\uparrow$ & Diversity$_k$ $\uparrow$ \\
     \midrule
    Real      & 1.023           & 0.25                  & 12.85 \\
    \midrule
    EDGE      & 2.876          & 0.14                  & 11.24 \\
    FineDance & 1.120            & 0.12                  & 11.39 \\
    \bottomrule
    \end{tabular}
    }
    \vspace{-0.1in}
    \caption{\small Evaluation of music-driven dance generation,  SMPL-H is adopted as the motion representation for training and testing. To enhance the diversity and accuracy of the model, we incorporate both our annotated AIST and FineDance~\cite{finedance} data for training and testing.
    }
    \label{tab:music2dance}
  \vspace{-0.1in}
\end{table}
To verify its usefulness in the 3D whole-body mesh recovery task, we use Hand4Whole~\cite{GyeongsikMoon2020hand4whole} as an example and evaluate MPVPE on the widely-used AGORA val~\cite{agora} and EHF~\cite{smpl-x} datasets. For the baseline model, we train it on the commonly used COCO~\cite{coco_wholebody}, Human3.6M~\cite{hm36}, and MPII~\cite{andriluka2014mpii} datasets. We then train another model by incorporating an additional $10\%$ of the single-view data sampled from \newdataname while keeping the other setting the same. As shown in Tab.~\ref{tab:mesh_recovery}, the model trained with \newdataname shows a significant decrease of $7.8\%$ in MPVPE on EHF and AGORA compared to the baseline model. The gains come from the increase in diverse appearances and poses in \newdataname, indicating the effectiveness and accuracy of the motion annotations in \newdataname and its ability to benefit the 3D reconstruction task. 
%
\begin{figure*}[t]
    \vspace{-0.1in}
    \begin{minipage}{0.99\textwidth}
        \centering
    \resizebox{\textwidth}{!}{
      \tiny
      \begin{tabular}{l|cc|cc|cc|cc|cc}
        \toprule
        \rowcolor{lightgray}
        \multirow{2}{*}{} & \multicolumn{2}{c|}{Body $\uparrow$} &
                           \multicolumn{2}{c|}{Foot $\uparrow$} &
                           \multicolumn{2}{c|}{Face $\uparrow$} &
                           \multicolumn{2}{c|}{Hand $\uparrow$} &
                           \multicolumn{2}{c}{Wholebody $\uparrow$} \\
        \rowcolor{lightgray}
        \multirow{-2}{*}{Method} & AP & AR & AP & AR & AP & AR & AP & AR & AP & AR \\
        \midrule
        DeepPose & 0.444 & 0.568 & 0.368 & 0.537 & 0.493 & 0.663 & 0.235 & 0.410 & 0.335 & 0.484 \\
        OpenPose & 0.563 & 0.612 & 0.532 & 0.645 & 0.765 & 0.840 & 0.386 & 0.433 & 0.442 & 0.523 \\
        SimpleBaseline   & 0.666 & 0.747 & 0.635 & 0.763 & 0.732 & 0.812 & 0.537 & 0.647 & 0.573 & 0.671 \\
        HRNet    & \textbf{0.701} & \textbf{0.773} & 0.586 & 0.692 & 0.727 & 0.783 & 0.516 & 0.604 & 0.586 & 0.674 \\
        PVT      & 0.673 & 0.761 & 0.660 & 0.794 & 0.745 & 0.822 & 0.545 & 0.654 & 0.589 & 0.689 \\
        FastPose50-dcn-si& 0.706 & 0.756 & \textbf{0.702} & 0.775 & 0.775 & 0.825 & 0.457 & 0.539 & 0.592 & 0.665 \\
        RTMPose-L& 0.695 & 0.769 & 0.658 & 0.785 & 0.833 & 0.887 & 0.519 & 0.628 & 0.616 & 0.700 \\
        RTMPose-L* (130 epoch) & 0.697 & 0.768 & 0.684 & \textbf{0.803} & \textbf{0.840} & 0.891 & 0.586 & 0.684 & 0.625 & 0.710 \\
        RTMPose-L* (270 epoch) & 0.699 & 0.768 & 0.684 & 0.803 & 0.839 & \textbf{0.892} & \textbf{0.594} & \textbf{0.691} & \textbf{0.629} & \textbf{0.712} \\
        \bottomrule
      \end{tabular}}
      \vspace{-0.1in}
      \captionof{table}{\small Quantitative results of 2D whole-body pose estimation on the COCO-Wholebody~\cite{coco_wholebody} dataset. Methods marked with an asterisk (*) indicate results obtained training on our data. The inclusion of \newdataname has significantly enhanced the keypoints training for RTMPose, particularly improving the accuracy of hand pose estimation.}
      \label{tab:rtmpose_table}
    \end{minipage}
  \hfill
  \begin{minipage}{0.99\textwidth}
    \vspace{0.03in}
      \centering
        \resizebox{\textwidth}{!}{
          \begin{tabular}[t]{c} \hspace{-4.6mm}
                \includegraphics[width=1\textwidth]{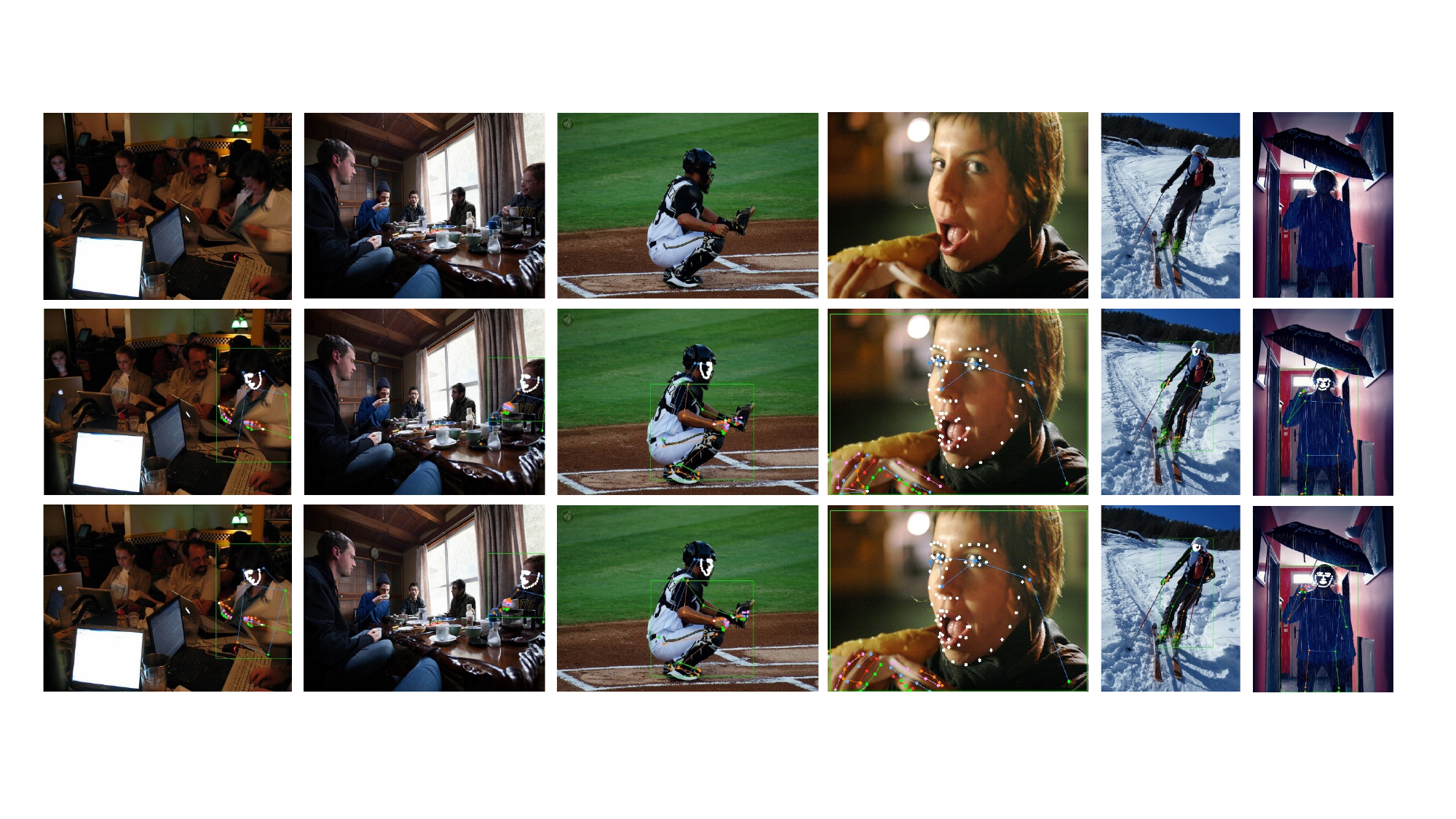}
            \end{tabular}
        }
      \vspace{-0.27in}
      \caption{\small Visualized results of 2D whole-body pose estimation on the COCO-Wholebody~\cite{coco_wholebody} dataset. Please zoom in for details.} 
      \vspace{-0.23in}
      \label{fig:rtmpose_fig}
  \end{minipage}
\end{figure*}

\subsubsection{Impact on 2d Whole-body Pose Estimation}
%
\noindent \textbf{Experiment Setup.} 
We use \newdataname subsets to train whole-body pose estimation, with ~\cite{rtmpose} as the baseline, and all models were trained at a uniform resolution of $256\times 192$.

\noindent \textbf{Evaluation metrics.} We use the AP (Average Precision) and AR (Average Recall) of the hand, face, body, and whole-body to evaluate the performance of 2D whole-body pose estimation trained on \newdataname.

\noindent \textbf{Evaluation results.} 
We set the ratio of different datasets as \newdataname  : COCO-Wholebody : Ubody $= 2:1:1.5$, with optimal downsampling rate 5 from ablation studies. Detailed results are shown in the Appendix Tab.~\ref{tab:rtmpose_supp}.
As illustrated in Tab.~\ref{tab:rtmpose_table} and Fig.~\ref{fig:rtmpose_fig}, we present the training results after adjusting the data proportions based on the ablation experiments. We follow training hyper-parameters, data proportions of RTMPose, and modify the learning rate decay strategy. 
Finally, the results on the COCO-Wholebody test set are one percentage higher than the baseline, achieving whole-body AP at 0.712 after 270 epochs of training.
\section{Conclusion}
In this paper, we introduce \newdataname, a large-scale, expressive, and precise multimodal whole-body human motion dataset. It overcomes the limitations of existing mocap datasets that primarily focus on indoor, body-only motions with limited action types. \newdataname includes 180.9 hours of whole-body motions, with corresponding text, audio, video, and annotated keypoints. We developed an automatic annotation pipeline to label 120.5K 3D motions, sequence-level semantic labels, and 19.5M frame-level pose descriptions. Experiments demonstrate the accuracy of our pipeline and the dataset's value in improving expressive and diverse motion generation, 3D human mesh recovery, and pose estimation.

\noindent\textbf{Limitation and future work.} Our markerless pipeline offers lower motion quality than multi-view, marker-based systems, and current evaluation metrics often misalign with visual results, necessitating refinement. \newdataname's multimodal scale can advance tasks like motion prior learning and multi-modality pre-training, especially with LLMs, and we expect it to drive further research.

\clearpage

\clearpage


\clearpage

\ifCLASSOPTIONcompsoc
  \section*{Acknowledgments}
\else
  \section*{Acknowledgment}
\fi


\ifCLASSOPTIONcaptionsoff
  \newpage
\fi



\bibliographystyle{IEEEtran}
\end{document}